%% file: main.tex
\documentclass[10pt,twocolumn,letterpaper]{article}

\usepackage{cvpr}              
\usepackage[accsupp]{axessibility}
\input{preamble}
\definecolor{cvprblue}{rgb}{0.21,0.49,0.74}
\usepackage[pagebackref,breaklinks,colorlinks,allcolors=cvprblue]{hyperref}
\usepackage{multirow}
\usepackage{array} 
\usepackage{siunitx,booktabs,adjustbox,tabularx}
\usepackage{threeparttable}
\usepackage{pifont}
\usepackage{diagbox}
\usepackage{xcolor,colortbl}
\newcommand{\cmark}{\textcolor{green!60!black}{\ding{51}}}
\newcommand{\xmark}{\textcolor{red!70!black}{\ding{55}}}
\usepackage{graphicx}
\usepackage{placeins}    
\usepackage{dblfloatfix}
\usepackage{makecell}
\usepackage{float}
\usepackage[ruled]{algorithm2e} 
\SetAlCapHSkip{0pt}

\SetAlgoCaptionLayout{algcapleft}
\SetAlgoCaptionSeparator{\space}
\SetKwFor{For}{\textbf{for}}{\ \textbf{do}}{\textbf{end for}}
\usepackage{sidecap}
\usepackage{lineno}

\def\paperID{} 
\def\confName{CVPR}
\def\confYear{2026}

\title{X-AVDT: Audio-Visual Cross-Attention for Robust Deepfake Detection}

\author{
    {Youngseo Kim}\quad
    {Kwan Yun}\quad
    {Seokhyeon Hong}\quad
    {Sihun Cha}\quad
    {Colette Suhjung Koo}\quad
    {Junyong Noh}\quad \vspace{0.1cm} \\ 
    Visual Media Lab, KAIST
}

\begin{document}
\maketitle
\input{sec/0_abstract}    
\input{sec/1_intro}
\input{sec/2_rw}
\input{sec/3_preliminaries}
\input{sec/4_method}
\input{sec/5_dataset}
\input{sec/6_experiments}
\input{sec/7_conclusion}
\section*{Acknowledgement}
{This work was supported by the Institute of Information \& Communications Technology Planning \& Evaluation (IITP) grant funded by the Korea government (MSIT) (No. RS-2024-00439499, Generating Hyper-Realistic to Extremely-stylized Face Avatar with Varied Speech Speed and Context-based Emotional Expression) (50\%), and by the Culture, Sports and Tourism R\&D Program through the Korea Creative Content Agency (KOCCA) grant funded by the Ministry of Culture, Sports and Tourism in 2024 (RS-2024-00440434) (50\%).}
{
    \small
    \bibliographystyle{ieeenat_fullname}
    \bibliography{main}
}
\appendix
\input{sec/X_suppl}
\end{document}

%% file: sec/0_abstract.tex
\begin{abstract}

The surge of highly realistic synthetic videos produced by contemporary generative systems has significantly increased the risk of malicious use, challenging both humans and existing detectors. Against this backdrop, we take a generator-side view and observe that internal cross-attention mechanisms in these models encode fine-grained speech–motion alignment, offering useful correspondence cues for forgery detection. Building on this insight, we propose \textbf{X-AVDT}, a robust and generalizable deepfake detector that probes generator-internal audio-visual signals accessed via DDIM inversion to expose these cues. X-AVDT extracts two complementary signals: (i) a video composite capturing inversion-induced discrepancies, and (ii) audio–visual cross-attention feature reflecting modality alignment enforced during generation. To enable faithful, cross-generator evaluation, we further introduce \textbf{MMDF}, a new multi-modal deepfake dataset spanning diverse manipulation types and rapidly evolving synthesis paradigms, including GANs, diffusion, and flow-matching. Extensive experiments demonstrate that X-AVDT achieves leading performance on MMDF and generalizes strongly to external benchmarks and unseen generators, outperforming existing methods with accuracy improved by \textbf{+13.1\%}. Our findings highlight the importance of leveraging internal audio–visual consistency cues for robustness to future generators in deepfake detection. Code is available at \href{https://youngseo0526.github.io/X-AVDT/}{X-AVDT}.
\vspace{-0.6cm}
\end{abstract}


%% file: sec/1_intro.tex
\section{Introduction}
Deepfakes, synthetic or edited portrait videos that manipulate identity, speech, or motion, have become feasible with advances in generative video models. These models have evolved from Generative Adversarial Networks (GANs)~\cite{goodfellow2014generative} to diffusion-based models~\cite{rombach2022high,ho2020denoising}, which push fidelity to unprecedented levels. Recent systems~\cite{cui2024hallo2,wei2024aniportrait,han2024face} can synthesize photorealistic digital humans from minimal input, lowering production costs for creative media and assistants. However, when misused, the same advances heighten societal and security risks, including targeted disinformation, real-time impersonation, identity theft, and financial fraud~\cite{Chesney2019,Sensity2019,WeProtect2024,FCC2024NHRobocall,Turing2024Behind,NIST2025DeepfakeEval}. These concerns have motivated the research community to develop deepfake detection, which aims to reliably authenticate media under rapidly evolving generators.
 
\input{figs/0_av_attn_map}

Diffusion models have become central to recent facial forgery generation. Many video generators employ cross-attention~\cite{vaswani2017attention} to condition visual features on external signals such as text, motion, or audio~\cite{bar2024lumiere,wang2024instantid,chen2025hunyuanvideo}. Among these external signals, audio is especially useful as it provides frame synchronous, densely informative supervision aligned with facial dynamics. In audio-driven diffusion models, this conditioning often takes the form of audio–visual cross-attention, which ties phonetic content to facial motion and expressiveness. Such architectures are explicitly designed to promote audio-visual alignment via cross-attention in the diffusion U-Net, making their internal features a natural source of correspondence cues for deepfake detection. Figure~\ref{fig:0_av_attn_map} presents examples of extracted audio-visual cross attention features, using Hallo~\cite{xu2024hallo}, an audio-driven talking head diffusion model pretrained on a large corpus of speech videos. It is observed that similar attention patterns recur across different generator frameworks. This indicates that internal audio–visual cross-attention features from diffusion models provide a robust, generator-agnostic discriminative signal for deepfake detection, reinforcing this generality.

\input{table/1_compare_dataset}

Building on this observation, we propose \textbf{X-AVDT}, an \textbf{A}udio-\textbf{V}isual \textbf{Cross}-Attention framework for robust \textbf{D}eepfake de\textbf{T}ection. Our framework leverages the cross-modal interaction of videos by utilizing fine-grained audio-visual diffusion features to generalize across manipulation types and synthesis models. To extract internal signals, we employ an inversion scheme that maps input videos into the diffusion model’s latent space and reconstructs them under the model prior. We further incorporate a latent noise map, reconstructed video, and input-reconstruction residual as complementary spatial cues, motivated by findings that pretrained diffusion models more faithfully reconstruct diffusion-generated content than real content~\cite{wang2023dire,cazenavette2024fakeinversion}. While fully synthetic videos often expose global inconsistencies, face-centric manipulations are confined to the facial region, preserve identity, and yield subtle artifacts that are easily obscured, thereby challenging residual-only detectors~\cite{wang2023dire,cazenavette2024fakeinversion}. By augmenting inversion-based discrepancies with audio-visual cross-attention and fusing them into a unified representation, X-AVDT provides complementary global and localized evidence that can improve detection reliability.

Existing datasets~\cite{li2019faceshifter,li2020celeb,dolhansky2020deepfake} are largely composed of earlier GAN-generated forgeries, offering limited coverage of contemporary models and manipulation types. As a result, they fail to capture the diversity and realism of continuously updated diffusion or flow-based methods, constraining progress toward building detectors that generalize beyond legacy benchmarks. To further facilitate robust detection of rapidly evolving deepfakes, we introduce \textbf{MMDF}, a curated \textbf{M}ulti-modal, \textbf{M}ulti-generator \textbf{D}eep\textbf{F}ake dataset. It is a high-quality dataset of paired real videos and corresponding fakes generated by a diverse suite of recent synthesis models. MMDF is the first dataset to cover contemporary diffusion (both U-Net \cite{rombach2022high} based and transformer \cite{peebles2023scalable} based) and flow-matching \cite{lipman2022flow} generators, and it includes audio–visual pairs. The dataset also spans manipulation paradigms such as talking-head generation \cite{cui2024hallo2,chen2025hunyuanvideo}, self-reenactment \cite{guo2024liveportrait,yang2025megactor}, and face swapping \cite{han2024face}, making it suitable for real-world, unconstrained deepfake detection. A comparison of MMDF with prior deepfake datasets is provided in Table~\ref{tab:1_compare_dataset}.

\input{figs/1_input}

%% file: figs/0_av_attn_map.tex
\begin{figure}
\includegraphics[width=\columnwidth]{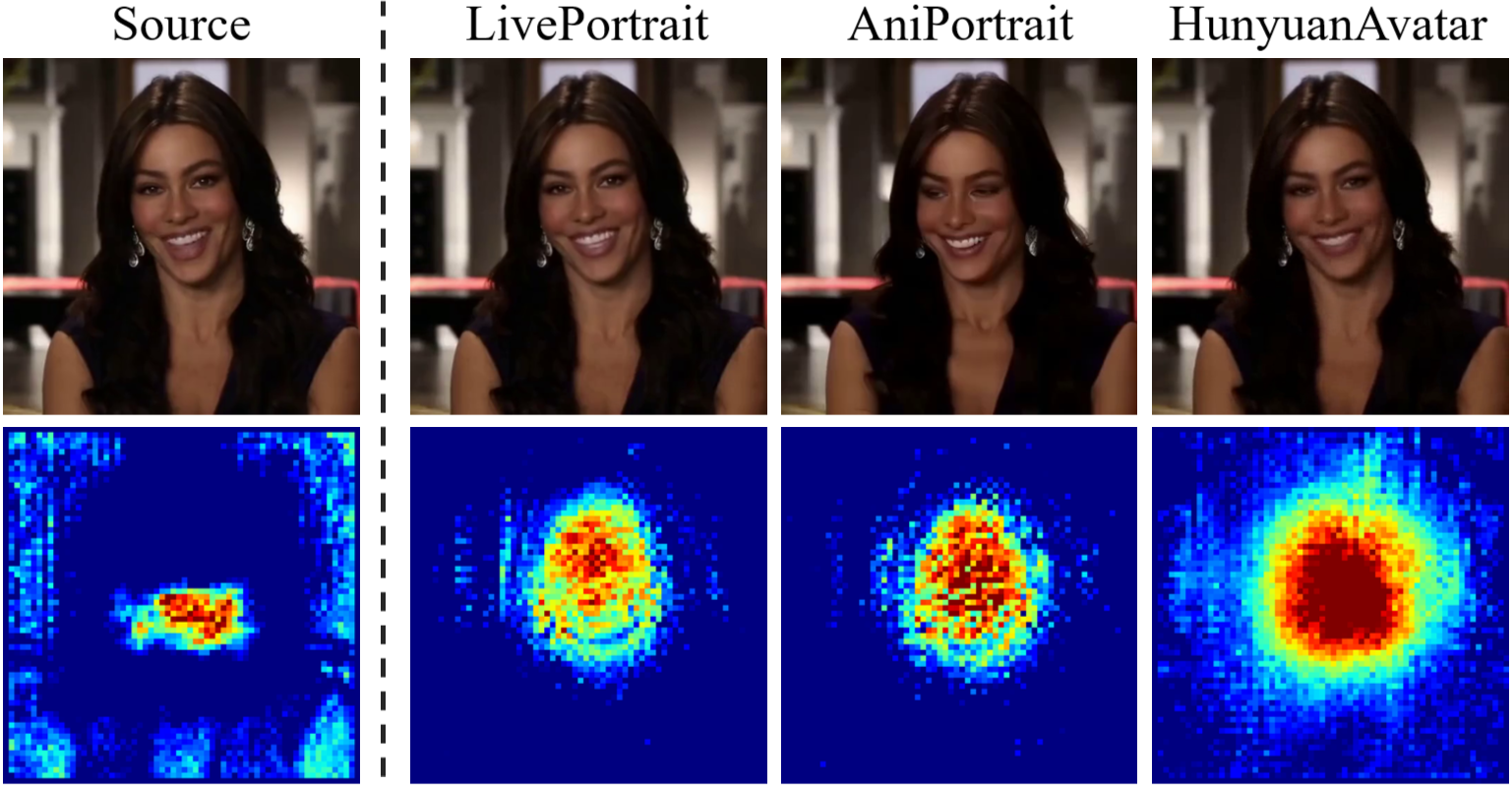}
\vspace{-0.6cm}
\caption{\textbf{Temporally averaged cross-attention maps.} For each video, we extract audio–visual cross-attention during DDIM inversion and average the maps over all frames to obtain a single heatmap. Real vs. fake samples exhibit consistent disparities.}
\vspace{-0.1cm}
\label{fig:0_av_attn_map}
\end{figure}

%% file: table/1_compare_dataset.tex
\begin{table}[t]
\centering
\footnotesize
\resizebox{\linewidth}{!}{
\begin{tabular}{l c ccc ccc}
\toprule
\multirow{2}{*}{Dataset} & \multirow{2}{*}{Modality} &
\multicolumn{3}{c}{Method} & \multicolumn{3}{c}{Model} \\
\cmidrule(lr){3-5}\cmidrule(lr){6-8}
& & {FS} & {RE} & {TH} & GAN & DF & FM \\
\midrule
Celeb-DF~\cite{li2020celeb} & V  & \cmark & \xmark & \xmark & \cmark & \xmark & \xmark \\
DF-Platter~\cite{narayan2023df} & V  & \cmark & \cmark & \xmark & \cmark & \xmark & \xmark \\
KoDF~\cite{kwon2021kodf} & AV  & \cmark & \cmark & \cmark & \cmark & \xmark & \xmark \\
FaceForensics++~\cite{rossler2019faceforensics++} & AV & \cmark & \cmark & \xmark & \cmark & \xmark & \xmark \\
DFDC~\cite{dolhansky2020deepfake} & AV & \cmark & \xmark & \xmark & \cmark & \xmark & \xmark \\
FakeAVCeleb~\cite{khalid2021fakeavceleb} & AV & \cmark & \xmark & \cmark & \cmark & \xmark & \xmark \\
AV-Deepfake1M~\cite{cai2024av} & AV & \xmark & \xmark & \cmark & \cmark & \xmark & \xmark \\
\rowcolor{black!6}
\textbf{MMDF} (\textbf{Ours}) & AV & \cmark & \cmark & \cmark & \cmark & \cmark & \cmark \\
\bottomrule
\end{tabular}
}
\vspace{-0.2cm}
\caption{\textbf{Comparison of deepfake video datasets.} FS:Face swapping, RE: Self-reenactment, TH: Talking-head generation, DF: Diffusion model, FM: Flow-matching.}
\vspace{-0.1cm}
\label{tab:1_compare_dataset}
\end{table}

%% file: figs/1_input.tex
\begin{figure*}[!t]
\centering 
\vspace{-0.7cm}
\includegraphics[width=\textwidth]{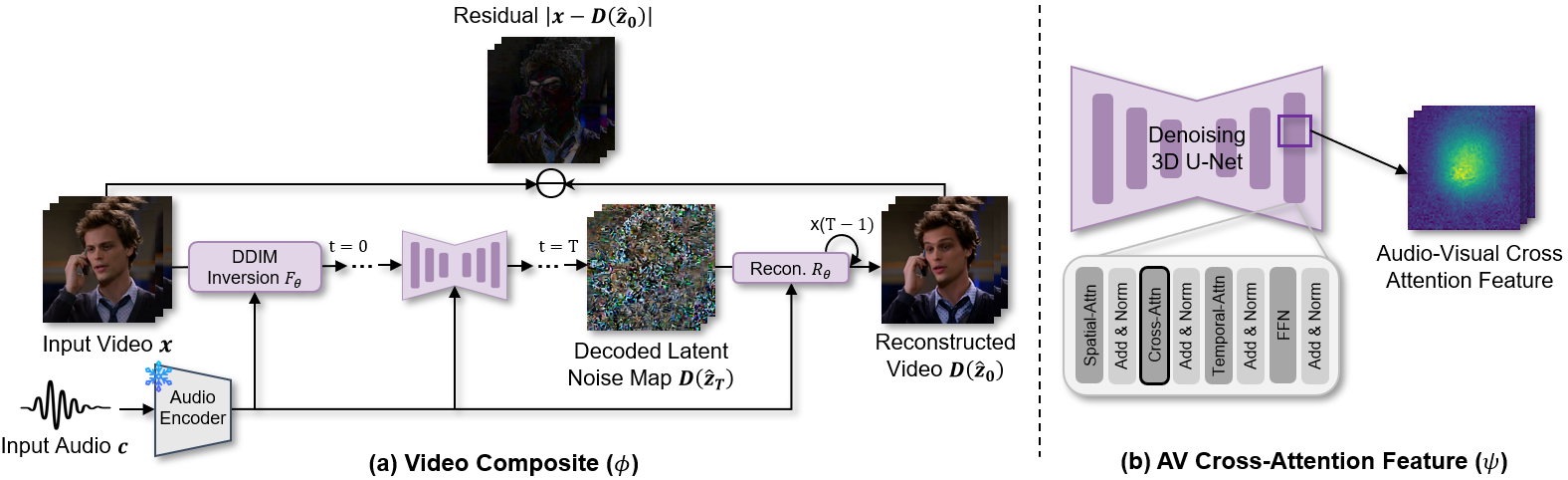}
\vspace{-0.5cm}
\caption{\textbf{Input representations with complementary features.} (a) video composite $\boldsymbol{\phi}$ is obtained from video $x$ and audio $c$ by running DDIM inversion and reconstruction, decoding both the noisy and clean latents, and computing the residual. We then concatenate four components channel-wise: $x$, $D(\hat z_T)$, $D(\hat z_0)$, and $\lvert x - D(\hat z_0)\rvert$. (b) AV cross-attention feature $\boldsymbol{\psi}$ is extracted during DDIM inversion from the diffusion U-Net and summarized as a frame-aligned tensor. These complementary cues (a) and (b) capture appearance information and modality alignment, respectively. For clarity, all visual elements shown ($D(\hat z_T)$, $D(\hat z_0)$, and $\lvert x - D(\hat z_0)\rvert$) are decoded images.}
\vspace{-0.1cm}
\label{fig:input}
\end{figure*}

%% file: sec/2_rw.tex
\section{Related Work}
\label{sec:rw}
\textbf{Deepfake Video Generation.} The synthesis and editing of human faces in video, with high realism and temporal coherence, have been extensively studied. Early innovations in generative modeling were driven by GANs~\cite{goodfellow2014generative} with gains in stability, resolution, and controllability~\cite{radford2015unsupervised,karras2017progressive,karras2019style}, along with basic conditional setups~\cite{mirza2014conditional,isola2017image,zhu2017unpaired}. More recently, diffusion and flow-matching models~\cite{rombach2022high,ho2020denoising,chen2025hunyuanvideo} have shown remarkable success in portrait video manipulation for deepfakes. In parallel, control signals such as audio~\cite{cui2024hallo2,wei2024aniportrait}, textual prompts~\cite{wu2023tune,bar2024lumiere}, facial landmarks~\cite{zhang2023adding, wang2024instantid}, dense motion flow~\cite{zhang2023adding}, and 3D face priors~\cite{poole2022dreamfusion, lin2023magic3d} have significantly enhanced the realism of synthetic videos. Moreover, current manipulation techniques include talking-head generation~\cite{cui2024hallo2,chen2025hunyuanvideo}, face reenactment~\cite{guo2024liveportrait,yang2025megactor}, face swapping~\cite{han2024face}, lip-sync/dubbing~\cite{prajwal2020lip,zhang2023sadtalker}, and appearance editing~\cite{brooks2022instructpix2pix}.

\noindent \textbf{Artifact-based Deepfake Detection.} Detection methods have evolved alongside deepfake video generation. Early attempts relied on hand-crafted forensic features, including blink detection~\cite{li2018ictu}, facial warping artifacts~\cite{li2018exposing}, or color inconsistencies~\cite{mccloskey2018detecting,li2020identification}. CNN-based classifiers trained on real and fake examples became dominant. Prior studies have demonstrated the effectiveness of deep learning for detecting subtle synthesis artifacts~\cite{rossler2019faceforensics++,afchar2018mesonet,li2020face,guera2018deepfake,qian2020thinking,shiohara2022detecting,sabir2019recurrent,wang2020cnn}. Frequency-domain and fingerprint-based detectors further improved generalization across different GAN architectures~\cite{frank2020leveraging, marra2019gans}. However, these methods often struggle to generalize beyond their training datasets and remain vulnerable to newer, more sophisticated forgeries.

\noindent \textbf{Generalizable Deepfake Detection.}
A principal goal is generalization across unseen generators, manipulation types, and domains. To this end, recent work investigates the semantic characteristics of generated images. DIRE~\cite{wang2023dire} proposes a reconstruction-based detector grounded in the hypothesis that images produced by diffusion models can be more accurately reconstructed by a pretrained diffusion model than real images. DRCT~\cite{chen2024drct} extends this idea by synthesizing hard examples via diffusion reconstruction and applying contrastive learning to the resulting residuals. FakeInversion~\cite{cazenavette2024fakeinversion} leverages features obtained via latent inversion of Stable Diffusion~\cite{rombach2022high}. These methods are built on prior observations that CLIP~\cite{radford2021learning} embeddings can be predictive of image authenticity~\cite{ojha2023towards,sha2023fake}.

\noindent \textbf{Audio-Visual Inconsistencies.} Several studies~\cite{zhou2021joint,feng2023self,smeu2025circumventing,klemt2025deepfake} adopt a late-fusion design in which RGB images and audio are encoded separately and combined only at the classification head. While this preserves strong per-modality features, the resulting embeddings occupy different latent spaces and are not directly aligned across modalities. Other approaches~\cite{haliassos2022leveraging,oorloff2024avff,liang2024speechforensics} 
learn audio–visual representations in a self-supervised way, implicitly fusing the modalities by pulling their embeddings together. Such implicit fusion can miss fine-grained semantic misalignment and offers limited interpretability of the cross-modal evidence. In contrast, we focus on internal features from large generative models as explicit, interpretable signals of persistent audio-visual inconsistency.

%% file: sec/3_preliminaries.tex
\section{Preliminaries}
\vspace{-0.1cm}
\input{figs/1_overview}
Diffusion models~\cite{dhariwal2021diffusion,ho2020denoising,rombach2022high} are probabilistic generative models that learn a data distribution by progressively denoising samples. They consist of two complementary processes. In the forward process, Gaussian noise is gradually injected into a clean image \(x_0\), given by
\begin{equation}
x_t \;=\; \sqrt{\bar{\alpha}_t}\,x_0 \;+\; \sqrt{1-\bar{\alpha}_t}\,\epsilon ,
\qquad \epsilon \sim \mathcal{N}(0,\mathbf{I})
\label{eq:fwd}
\end{equation}
which maps \(x_0\) to increasingly noisy latents \(x_t\), where \(t=0,\ldots,T\) and \(\bar{\alpha}_t=\prod_{k=1}^{t}\alpha_k\).
The reverse process starts from a noisy sample \(x_T\) and iteratively produces
cleaner states conditioned on an auxiliary vector \(c\) (e.g., audio),
using a noise predictor \(\epsilon_\theta(x_t,t,c)\).
Given the current estimate of the clean image,
\begin{equation}
\hat x_0(x_t,t,c)
=\frac{x_t-\sqrt{1-\bar\alpha_t}\,\epsilon_\theta(x_t,t,c)}{\sqrt{\bar\alpha_t}},
\end{equation}
the conditional reverse update can be written as
\begin{equation}
x_{t-1}
=\sqrt{\bar\alpha_{t-1}}\,\hat x_0(x_t,t,c)
+\sqrt{1-\bar\alpha_{t-1}}\,\epsilon_\theta(x_t,t,c).
\end{equation}

In this work, we rely on a pre-trained audio-conditioned Latent Diffusion Model (LDM)~\cite{rombach2022high}, where the diffusion process operates in the latent space of a VAE: $z_0 = E(x_0)$ and $x_0 \approx D(z_0)$. The denoiser is a 3D U-Net conditioned on external signals and composed of residual blocks, spatial self-attention, temporal self-attention, and cross-attention. Let $f^l_n$ be the video features at layer $l$ and video frame $n$, projected to queries $q_n^l$, keys $k_n^l$, and values $v_n^l$. The attention output is given by
\begin{equation}
\tilde{f}^{\,l}_n \;=\; A_n^l\, v_n^l,
\qquad
A_n^l \;=\; \mathrm{Softmax}\!\left(\frac{q_n^l (k_n^l)^{\!\top}}{\sqrt{d}}\right),
\label{eq:selfattn}
\end{equation}
where $d$ is the query/key embedding dimension. This enables the model to capture global dependencies across space and time in the video. In parallel, cross-attention is computed between video queries and the conditioning audio embedding, thereby guiding the denoising process toward the target reconstruction.

%% file: figs/1_overview.tex
\begin{figure*}
\centering
\vspace{-0.7cm}
\includegraphics[width=0.8\textwidth]{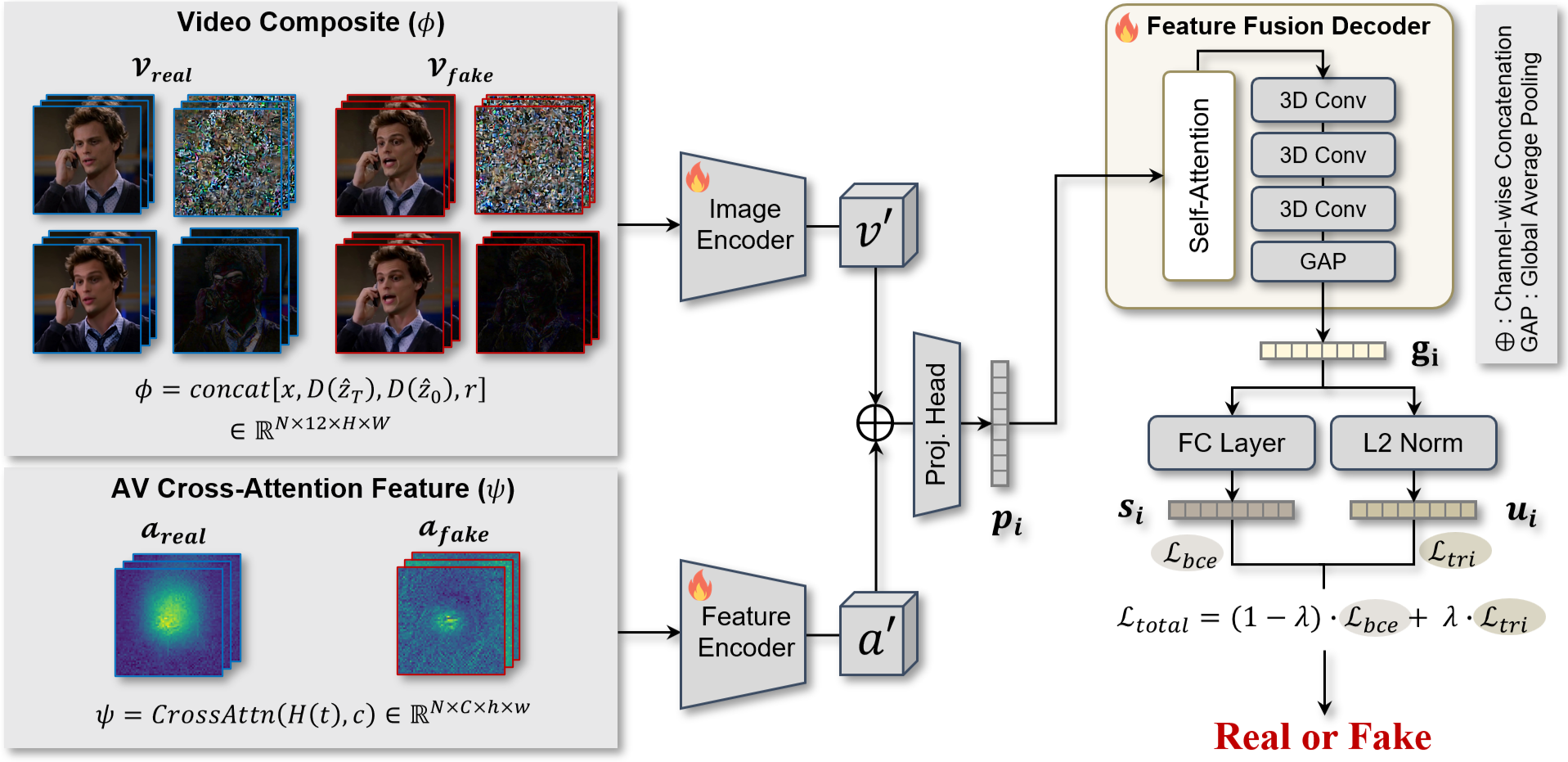}
\vspace{-0.1cm}
\caption{\textbf{The overall framework of X-AVDT.} From each audio-visual pair, we form two inputs $\boldsymbol{\phi}$ and $\boldsymbol{\psi}$. Two 3D encoders map them to features that are concatenated and passed through the Feature Fusion Decoder to produce a fused feature. A classification head outputs the real/fake score, while an embedding head is trained with a triplet objective to improve robustness.}
\vspace{-0.1cm}
\label{fig:overview}
\end{figure*}

%% file: sec/4_method.tex
\section{Method}
\subsection{Problem Definition}
We train a binary deepfake detector on audio–visual pairs $\mathcal{D}_{\text{train}}=\{(x_m,c_m,y_m)\}_{m=1}^{M}$,
where each example consists of a face video clip $x_m$, its paired audio condition $c_m$, and a label $y_m\in\{0,1\}$ that indicates whether the video is real or fake. The training set contains $M$ videos. From each input pair, an audio-conditioned LDM extracts two types of features, video composite $\boldsymbol{\phi}(x,c)$ and an
AV cross-attention feature $\boldsymbol{\psi}(x,c)$, as illustrated in Figure~\ref{fig:input} and discussed in Sec.~\ref{sec:input}. We fuse $\boldsymbol{\phi}$ and $\boldsymbol{\psi}$ and train a detector $G_\theta$ to estimate $p_\theta(y=1\mid x,c)=\sigma(G_\theta(\cdot))$ by optimizing a weighted sum of a binary cross-entropy term and a metric-learning term, which will be defined in Sec.~\ref{sec:arch}.

\subsection{Input Representation}
\label{sec:input}
\subsubsection{Diffusion Inversion \& Reconstruction}
As shown in Figure~\ref{fig:input}a, given a video $x\in\mathbb{R}^{N\times 3\times H\times W}$, where $N$ denotes the number of frames in the clip and $H$ and $W$ denote the height and width, respectively, and a paired audio condition $c$, we operate in the VAE latent space with encoder $E$ and decoder $D$. We first encode the input video frame to $z_0=E(x)$ while conditioning on the wav2vec 2.0~\cite{baevski2020wav2vec} audio embedding c (omitted in the figure for brevity). We then obtain the corresponding latent noise map using the DDIM inversion process $\hat z_T = F_\theta(z_0, c)$. Subsequently, starting from $\hat z_T$, we run the conditional reverse diffusion to obtain a clean latent $\hat z_0 = R_\theta(\hat z_T, c)$. We then decode $\hat z_0$ to the pixel space to obtain the reconstructed video $D(\hat z_0)$ and compute the residual between the input and the reconstruction $r=\lvert x-D(\hat z_0)\rvert$.

To construct the video composite $\boldsymbol{\phi}$ for the detector, we use an input formed by channel-wise concatenation of the image $x$, the decoded latent DDIM noise map $D(\hat z_T)$, the image $D(\hat z_0)$ reconstructed from the reverse DDIM process, and the reconstruction residual $r=\lvert x-D(\hat z_0)\rvert$:
\begin{equation}
\boldsymbol{\phi}(x,c)=\mathrm{concat}\big[\,x,\; D(\hat z_T),\; D(\hat z_0),\; r\,\big]\in\mathbb{R}^{N\times 12\times H\times W},
\end{equation}
Because DDIM inversion uses a finite number of steps, the mismatch after a forward-reverse pass reflects discretization error, whereas manipulated samples tend to produce smaller discrepancies and thus higher likelihood of forgery under the diffusion model~\cite{cazenavette2024fakeinversion}. We use the pattern of this gap as an inversion-induced discrepancy measure for detecting manipulation.

\subsubsection{Audio-Visual Cross-Attention Feature}
As illustrated in Figure~\ref{fig:input}b, DDIM inversion is performed with a 3D U-Net composed of multiple blocks organized into down, mid and up stages. Each block contains an audio-visual cross-attention layer in which video hidden states serve as queries while the audio encoder’s hidden states provide keys and values. We extract the cross-attention from an up block at a chosen diffusion timestep $t$, conditioned on the input audio. This attention is taken from the same conditioned LDM used to construct the video composite $\boldsymbol{\phi}$.

Let $H(t)$ be the 3D U-Net hidden state at timestep $t$. Using the attention output projection, we aggregate the multi-head outputs, reduce the head dimension to $C$ channels, and reshape the result into the per-frame latent grid, yielding a temporally aligned feature $\boldsymbol{\psi}(x,c)$, which is defined as:
\begin{equation}
\boldsymbol{\psi}(x,c)
=\mathrm{CrossAttn}\!\left(H(t),\,c\right)\in\mathbb{R}^{N\times C\times h\times w},
\end{equation}
Here, $h \times w$ is the latent-space resolution (e.g., $64 \times 64$ for $512 \times 512$ inputs with an 8x downsampling). In our setup, we extract the attention from the last up block at timestep $t=24$ and reshape it to a per-frame. The resulting $\boldsymbol{\psi}$ provides a compact, temporally aligned descriptor of audio–visual correspondence that the detector can exploit to distinguish authentic from manipulated content. Because it captures speech–motion synchrony enforced by the denoiser instead of appearance alone, it is less sensitive to purely visual artifacts and thus offers a complementary, model-internal cue that improves robustness. Ablation on input representations and attention features appear in Sec.~\ref{sec:abl}.

\subsection{Detector Architecture}
\label{sec:arch}
Figure~\ref{fig:overview} presents the X-AVDT framework. X-AVDT pairs two signals extracted from different parts of the conditioned LDM pipeline to improve robustness. Accordingly, the detector takes two inputs. Two 3D encoders, $E_v$ for $\boldsymbol{\phi}$ and $E_a$ for $\boldsymbol{\psi}$, produce feature volumes that are aligned in space and time and then fused by a Feature Fusion Decoder (FFD) to yield a logit and an embedding:
\vspace{-0.1cm}
\begin{equation}
\mathbf{v}'=E_v(\boldsymbol{\phi}), \qquad \mathbf{a}'=E_a(\boldsymbol{\psi}).
\end{equation}
The tensors $\mathbf{v}'$ and $\mathbf{a}'$ are concatenated along the channel dimension and projected to a shared embedding with a $1{\times}1$ convolution to obtain $\mathbf{p}_i$. The FFD then applies a self-attention layer over spatial tokens, followed by a series of $L$ 3D ResNeXt~\cite{xie2017aggregated} layers. Then global average pooling (GAP) produces a fused feature vector $\mathbf{g}_i$ for the $i$-th sample:
\vspace{-0.3cm}
\begin{align}
\label{eq:ffd}
\mathbf{p}_i &= \mathrm{Proj}\!\left(\mathrm{concat}\!\left[\mathbf{v}'_i,\mathbf{a}'_i\right]\right),\\
\mathbf{g}_i &= \mathrm{GAP}\!\left(\mathrm{Conv3D}^{L}\!\big(\mathrm{SelfAttn}(\mathbf{p}_i)\big)\right).
\end{align}
\vspace{-0.3cm}

From $\mathbf{g}_i$ we form two branches: a fully connected layer maps $\mathbf{g}_i$ to a scalar logit $s_i$ for the binary classification loss, and an embedding head outputs an $\ell_2$-normalized vector $\mathbf{u}^{(i)}$ for metric learning. With ground-truth labels $y_i\in\{0,1\}$ (real=0, fake=1) and sigmoid $\sigma(\cdot)$, the binary cross-entropy loss over a mini-batch of size $B$ is defined as follows:
\vspace{-0.3cm}
\begin{equation}
\mathcal{L}_{\text{bce}} =
-\frac{1}{B} \sum_{i=1}^{B}
\Big[
y_i \log \sigma(s_i)
+ (1 - y_i)\log \bigl(1 - \sigma(s_i)\bigr)
\Big].
\end{equation}

Let $\mathbf{u}_a^{(i)}$, $\mathbf{u}_p^{(i)}$, and $\mathbf{u}_n^{(i)}$ denote anchor, positive, and negative embeddings for the $i$-th triplet, respectively. It tightens same class embeddings and separates different classes. Using the squared $\ell_2$ distance and a margin $m>0$, the triplet loss is
\vspace{-0.3cm}
\newcommand{\dist}[2]{\left\| #1 - #2 \right\|_2^2}
\begin{equation}
\label{eq:tri}
\mathcal{L}_{\text{tri}}
= \frac{1}{B}\sum_{i=1}^{B}
\max\bigl(0,\;
\dist{\mathbf{u}_a^{(i)}}{\mathbf{u}_p^{(i)}}
-\dist{\mathbf{u}_a^{(i)}}{\mathbf{u}_n^{(i)}} + m \bigr).
\end{equation}

The overall objective minimizes a weighted sum of the two terms, controlled by a balancing parameter $\lambda\in[0,1]$:
\begin{equation}
\label{eq:obj}
\mathcal{L}_{\text{total}}
=(1-\lambda)\,\mathcal{L}_{\text{bce}}
+\lambda\,\mathcal{L}_{\text{tri}}.
\end{equation}
This joint objective allows the model to learn discriminative and mutually informative representations that generalize across manipulation patterns. Further analysis is provided in Sec.~\ref{sec:abl}.

%% file: sec/5_dataset.tex
\section{Dataset}
The MMDF dataset was constructed from curated real clips paired with their corresponding manipulated versions, resulting in a collection of 28.8k clips with a total duration of 41.67 hours. Using Hallo3 dataset~\cite{cui2025hallo3} as our source, we applied each video clip duration, resolution, and face-presence filters with MediaPipe~\cite{lugaresi2019mediapipe} to retain single-person, frontal-to-quarter views with stable lip motion and speech. We removed clips with scene cuts, strong camera motion, excessive facial motion, or long side facing poses. These videos contain individuals of diverse ages, ethnicities, and genders in close-up shots across varied indoor and outdoor backgrounds. MMDF covers three manipulation types, such as talking-head generation, self-reenactment, and face swapping, produced using GAN, diffusion, and flow-matching generators. Comparative networks were trained on a diverse set of data obtained from commonly used generators and evaluated on unseen ones to assess cross-generator generalization. See Table~\ref{tab:2_dataset} for details.

\input{table/2_dataset}
\input{table/3_dataset_quality}

To quantify the realism of the manipulated videos, we report several metrics values. As shown in Table~\ref{tab:3_dataset_quality}, MMDF exhibits stronger audio–visual synchronization and more favorable perceptual and video statistics than FaceForensics++~\cite{rossler2019faceforensics++} and FakeAVCeleb~\cite{khalid2021fakeavceleb}, indicating temporally coherent, high-quality manipulations suitable for the detection task. In addition, we report Human False Acceptance Rate (HFAR), defined as the proportion of fake clips judged to be real by human evaluators (i.e., False Positive Rate, FPR). A higher HFAR indicates that humans are more likely to perceive fake clips as real, suggesting that the dataset contains more challenging and realistic negative samples. Metric definitions are provided in Sec.~\ref{sec:imp} and MMDF details are in the supplementary material.

%% file: table/2_dataset.tex
\begin{table}[t]
\centering
\footnotesize
\resizebox{\columnwidth}{!}{
\begin{tabular}{
    >{\centering\arraybackslash}m{0.8cm}
    >{\centering\arraybackslash}m{2.3cm}
    >{\centering\arraybackslash}m{2.6cm}
    >{\centering\arraybackslash}m{2.1cm}
    >{\centering\arraybackslash}m{1cm}
}
\toprule
Split & Generator & Model & Method & \#Clips (real/fake) \\
\midrule
\multirow{3}{*}{Train}
    & Hallo2 \cite{cui2024hallo2} & Diffusion & Talking-Head & 4k/4k \\
    & LivePortrait \cite{guo2024liveportrait} & GAN & Self-Reenactment & 4k/4k \\
    & FaceAdapter \cite{han2024face} & Diffusion & Face Swapping & 4k/4k \\
\midrule
\multirow{3}{*}{Test}
    & HunyuanAvatar \cite{chen2025hunyuanvideo} & Flow Matching  & Talking-Head & 0.8k/0.8k \\
    & MegActor-$\Sigma$ \cite{yang2025megactor} & Diffusion Transformer & Self-Reenactment & 0.8k/0.8k \\
    & AniPortrait \cite{wei2024aniportrait} & Diffusion  & Talking-Head & 0.8k/0.8k \\
\bottomrule
\end{tabular}}
\vspace{-0.2cm}
\caption{\textbf{Composition of the dataset.} This coverage better reflects current synthesis trends than outdated datasets that focus primarily on GANs, and permits assessing cross-generator generalization.}
\label{tab:2_dataset}
\end{table}

%% file: table/3_dataset_quality.tex
\begin{table}[t]
\centering
\footnotesize
\resizebox{\linewidth}{!}{
\begin{tabular}{lccccc}
\toprule
Dataset & Sync-C $\uparrow$ & Sync-D $\downarrow$ & LPIPS $\downarrow$ & FVD $\downarrow$ & HFAR $\uparrow$ \\
\midrule
FaceForensics++~\cite{rossler2019faceforensics++}    & 3.32 & 11.06 & 0.27 & 370.23 & 0.22 \\
FakeAVCeleb~\cite{khalid2021fakeavceleb}  & 5.87 & 8.38 & 0.19 & 170.61 & 0.34 \\
\midrule
\textbf{MMDF} (\textbf{Ours})  & \textbf{7.36} & \textbf{7.35} & \textbf{0.07} & \textbf{121.39} & \textbf{0.41} \\
\bottomrule
\end{tabular}
}
\vspace{-0.2cm}
\caption{\textbf{Audio-visual quality of manipulated videos.} Sync-C, Sync-D, LPIPS, and FVD are computational metrics, whereas HFAR denotes the human false acceptance rate measured in a user study.}
\vspace{-0.3cm}
\label{tab:3_dataset_quality}
\end{table}

%% file: sec/6_experiments.tex
\input{table/4_comparison}
\input{table/5_comparison_benchmark}
\section{Experiments}
\subsection{Implementation Details}
\label{sec:imp}
X-AVDT is implemented and trained with the same configuration across all evaluation datasets. Training requires 14 hours on a single NVIDIA RTX 3090 GPU.

\noindent\textbf{Architecture.} Hallo~\cite{xu2024hallo} is employed as our audio-conditioned latent diffusion backbone, initialized from Stable Diffusion~\cite{rombach2022high}. While we choose Hallo because it provides high-fidelity synthesis and rich internal audio–visual signals that our detector can exploit, other choices can also be possible~\cite{liu2025moee,wei2024aniportrait,chen2025echomimic}. See the supplementary material for details. Audio conditioning is provided by wav2vec 2.0~\cite{baevski2020wav2vec} features projected to the U-Net cross-attention embedding dimension. We do not use classifier-free guidance during either DDIM inversion or reconstruction to preserve bijectivity and conditioning fidelity. Unless otherwise stated, we extract the audio–visual cross-attention from the last up block of the U-Net at diffusion timestep $t=24$. Both the image encoder $E_v(\cdot)$ and the feature encoder $E_a(\cdot)$ use 3D ResNeXt~\cite{xie2017aggregated}. The FFD uses an $L$-layer 3D ResNeXt stack. We fix $L=3$ in Eq. (\ref{eq:ffd}). In our setup, the video composite $\boldsymbol{\phi}$ has 12 channels ($x$, $D(\hat z_T)$, $D(\hat z_0)$, residual), and the AV cross-attention feature $\boldsymbol{\psi}$ uses $C=320$ channels after collapsing the multi-head dimension and is reshaped with a latent resolution of $64\times64$.

\noindent\textbf{Training Details.} We train the detector for 2 epochs on frames of size $512\times512$ using AdamW~\cite{loshchilov2017decoupled} with a learning rate of $1\times10^{-4}$, weight decay of $0.05$, and a batch size of $8$. For the triplet loss Eq. (\ref{eq:tri}), we set the margin $m$ to $0.3$, and the balancing parameter $\lambda$ in the overall objective Eq. (\ref{eq:obj}) is also set to $0.3$.

\vspace{3mm}
\noindent\textbf{Automatic Metrics.} We report the values measured by four detection metrics, AUROC, Average Precision (AP), Accuracy at Equal Error Rate (Acc@EER), and Accuracy. We obtain Acc@EER by computing the EER threshold from the ROC curve and evaluating accuracy at that threshold. To assess the proposed MMDF dataset, we also report values measured by generation quality metrics including Lip Sync scores (Sync-C and Sync-D) from SyncNet for lip-speech alignment~\cite{chung2016out}, LPIPS for perceptual distance~\cite{zhang2018unreasonable}, and Fr\'echet Video Distance (FVD) for distributional distance between generated and ground-truth videos~\cite{unterthiner2018towards}. Lower is better for LPIPS and FVD, higher Sync-C and lower Sync-D indicate better synchronization. 

\input{table/6_ablation_attn}

\noindent\textbf{Baselines.} We evaluated X-AVDT against open-source baselines, including the video-only LipForensics~\cite{haliassos2021lips}, and audio–visual methods RealForensics~\cite{haliassos2022leveraging}, AVAD~\cite{feng2023self}, LipFD~\cite{liu2024lips}, FACTOR~\cite{reiss2023detecting}, and AVH-Align~\cite{smeu2025circumventing}.  We used two variants: (i) the official checkpoint and (ii) models retrained from scratch on our training sets using the official code. This protocol enables a fair cross-dataset transfer comparison against our MMDF-trained model. Further details are in the supplementary material.

\subsection{Comparison with State-of-the-art}
\subsubsection{Comparison on MMDF Dataset}
Table~\ref{tab:4_mmdf_comp} comprises two panels. \textit{Official-pretrained} reports results from publicly released checkpoints of prior methods, and \textit{MMDF-retrained} reports the same baselines retrained on our MMDF for cross-dataset evaluation. Baselines are evaluated in both settings, whereas X-AVDT values are reported in the \textit{MMDF-retrained} panel. As the training code for LipForensics~\cite{liu2024lips} and AVAD~\cite{feng2023self} is unavailable, we report results obtained by their official pretrained model only. The pretrained baselines presented in the first panel tended to overfit to earlier synthesis methods and showed limited domain adaptation capability, resulting in poor transfer ability to unseen manipulations and a persistent generalization gap across datasets. As shown in the bottom row, X-AVDT achieved the highest average AUROC of 95.29, exceeding the performance of the strongest retrained baseline (RealForensics~\cite{haliassos2022leveraging} 92.42). The margin over prior methods remained broadly consistent across generators, resulting in a clear overall lead despite MMDF retraining.

\subsubsection{Comparison on Benchmark Dataset}
We further evaluate X-AVDT on the GAN-based benchmarks FakeAVCeleb~\cite{khalid2021fakeavceleb} and FaceForensics++~\cite{rossler2019faceforensics++}, following the same setup as in Table~\ref{tab:4_mmdf_comp}. As summarized in Tables~\ref{tab:4_mmdf_comp} and \ref{tab:5_ben_comp}, models trained on MMDF transfer well to these benchmarks. Notably, several pretrained detectors are marked with \textsuperscript{\dag}, indicating train-test overlap. Even under this favorable condition for the baselines, X-AVDT achieved the best scores on both benchmarks with an AUROC of 99.69 on FakeAVCeleb and 89.55 on FaceForensics++.

\input{figs/2_ablation_attn}
\input{figs/5_perturbation}
\input{figs/4_grad_cam}

\subsection{Human Evaluation}
For human evaluation studies, we use two metrics: Human Evaluation (HE) in Table~\ref{tab:4_mmdf_comp} and Table~\ref{tab:5_ben_comp}, and HFAR in Table~\ref{tab:3_dataset_quality}. Both HE and HFAR were obtained from the same user study with 24 participants (balanced in gender, aged between their 20s and 30s), where each participant viewed the videos with audio and answered whether it was real or fake. Across datasets, HE was consistently lower than our detector, indicating the task is challenging for humans, whereas our model remains robust.
\input{table/7_ablation_input_loss}
 
\subsection{Ablation Study}
\label{sec:abl}
\noindent\textbf{Choice of Attention Type and Timestep.} We analyze how the choice of attention feature and the diffusion timestep affect detection. Table~\ref{tab:6_ablation_attn_feat} shows that, among attention features extracted from the 3D U-Net in Hallo~\cite{xu2024hallo}, the audio-conditioned cross-attention was consistently the most informative, as it explicitly regularizes the representation toward audio–visual alignment. Figure~\ref{fig:abl_attn} further illustrates that temporal coherence and spatial appearance tend to capture global motion and pose changes, making them less discriminative. By contrast, cross-attention highlights articulators while suppressing background and is inherently less sensitive to scene changes. Moreover, features from earlier diffusion steps were more discriminative than those from later steps, because early denoising retains stronger conditioning signals before texture refinement dominates, leaving modality-consistency cues less degraded.

\noindent\textbf{Complementary Effect of Input Representations.} The video composite $\boldsymbol{\phi}$ encodes inversion-induced discrepancies, while the AV cross-attention feature $\boldsymbol{\psi}$ provides cross-modal consistency cues derived from the diffusion model’s internal alignment. Motivated by their complementary nature, we combine $\boldsymbol{\phi}$ and $\boldsymbol{\psi}$ and assess their contributions via an ablation study that removes each component individually. The combined model yields the highest overall performance, indicating that the two representations reinforce each other. Table~\ref{tab:7_abl_input_loss}a corroborates this with quantitative results across all three metrics.

\noindent\textbf{Effect of Loss Design.} To assess the effect of loss formulation, we compared models trained with only the binary cross-entropy loss $\mathcal{L}_{\text{bce}}$ against those using the combined objective $\mathcal{L}_{\text{bce}} + \mathcal{L}_{\text{tri}}$. The triplet loss $\mathcal{L}_{\text{tri}}$ serves as an auxiliary metric-learning loss, providing an inductive bias toward a more discriminative class structure in the embedding space that complements the classification signal. See Table~\ref{tab:7_abl_input_loss}b for quantitative results.

\noindent\textbf{Robustness of In-the-Wild Perturbation Attack.} We trained X-AVDT on MMDF without augmentation and evaluated robustness on unseen corruptions with five severity levels. For comparison, all baselines used their official pretrained checkpoints. We report AUROC (\%) values in Figure~\ref{fig:5_abl_perturbation}. X-AVDT outperformed prior detectors across severity levels, with smaller performance decay than comparative methods under high-frequency suppressing distortions (JPEG/Blur), additive noise, and scale changes due to resizing, as well as under temporal disruptions from frame dropping. Detailed information can be found in the supplementary material.

\subsection{Discussion}
\vspace{-0.1cm}
To understand how our detector treats real versus forged videos, we applied Grad-CAM~\cite{selvaraju2017grad} to our detector and visualized activation maps on samples from three representative generators (Figure~\ref{fig:grad_cam}). Grad-CAM highlights where the model grounds its decision in each frame, and these activation maps make the effect of our design explicit. Our method concentrated activation on articulatory regions and maintained this focus across frames in real videos. In contrast, forgeries elicited scattered, multi-focal responses, revealing the absence of consistent audio–visual cues. This behavior was consistent across all generators, indicating that our detector exploits audio–visual consistency as a general cue, which in turn explains its generalization beyond any single model’s artifacts. See the supplementary material for details.
\vspace{-0.1cm}

%% file: table/4_comparison.tex
\begin{table*}[ht]
\vspace{-0.5cm}
\centering
\setlength{\tabcolsep}{2pt}  
\resizebox{\linewidth}{!}{
\begin{tabular}{@{}>{\raggedright\arraybackslash}m{2.8cm}
  *4{>{\centering\arraybackslash}m{1.2cm}}
  *4{>{\centering\arraybackslash}m{1.2cm}}
  *4{>{\centering\arraybackslash}m{1.2cm}}
  *4{>{\centering\arraybackslash}m{1.2cm}}
}
\toprule
& \multicolumn{4}{c}{AniPortrait~\cite{wei2024aniportrait}} & \multicolumn{4}{c}{MegActor-$\Sigma$~\cite{yang2025megactor}} & \multicolumn{4}{c}{HunyuanAvatar~\cite{chen2025hunyuanvideo}} & \multicolumn{4}{c}{Average} \\
\cmidrule(lr){2-5}\cmidrule(lr){6-9}\cmidrule(lr){10-13}\cmidrule(lr){14-17}
Model & AUROC & AP & Acc@EER & \hspace{5pt}Acc & AUROC & AP & Acc@EER & \hspace{5pt}Acc & AUROC & AP & Acc@EER & \hspace{5pt}Acc & AUROC & AP & Acc@EER & \hspace{5pt}Acc \\
\midrule
\midrule
\multicolumn{17}{l}{\textit{Official-pretrained}} \\
\midrule
LipForensics~\cite{haliassos2021lips} & 79.49 & 81.45 & 73.25 & 74.50 & 72.17 & 73.56 & 65.85 & 67.11 & 63.66 & 56.50 & 59.91 & 60.66 & 74.24 & 74.54 & 71.91 & 72.38 \\
RealForensics~\cite{haliassos2022leveraging} & 86.31 & 85.23 & 78.65 & 77.56 & 62.93 & 61.60 & 58.79 & 58.71 & 64.90 & 60.30 & 61.40 & 60.85 & 77.65 & 75.71 & 72.15 & 65.70  \\
AVAD~\cite{feng2023self} & 73.25 & 74.98 & 66.05 & 67.49 & 50.87 & 56.96 & 51.26 & 51.18 & 75.37 & \underline{74.54} & \underline{74.14} & \underline{74.39} & 67.47 & 67.29 & 64.75 & 65.29  \\
FACTOR~\cite{reiss2023detecting} & 94.21 & 93.54 & 86.29 & 86.55 & 75.44 & 72.83 & 68.82 & 69.33 & 36.69 & 46.18 & 38.27 & 56.15 & 68.78 & 70.85 & 64.46 & 70.68 \\
LipFD~\cite{liu2024lips} & 52.55 & 50.60 & 51.74 & 51.90 & 51.45 & 49.89 & 50.80 & 50.93 & 60.38 & 53.85 & 57.78 & 58.30 & 57.64 & 51.83 & 55.37 & 55.82 \\
AVH-Align~\cite{smeu2025circumventing} & 74.88 & 73.94 & 68.82 & 70.03 & 51.20 & 47.81 & 50.86 & 50.93 & 34.53 & 39.96 & 37.50 & 36.91 & 50.34 & 50.95 & 49.89 & 49.98 \\
\midrule
\multicolumn{17}{l}{\textit{MMDF-retrained (cross-dataset evaluation)}} \\
\midrule
RealForensics~\cite{haliassos2022leveraging} & 97.47 & 97.22 & 90.99 & 90.28 & \textbf{97.60} & \textbf{96.70} & \textbf{92.68} & \textbf{92.45} & 74.89 & 72.09 & 68.38 & 61.12 & \underline{92.42} & \underline{91.39} & \underline{84.01} & \underline{81.28} \\
LipFD~\cite{liu2024lips} & 49.59 & 50.26 & 49.79 & 49.75 & 50.32 & 50.82 & 50.14 & 50.18 & 48.39 & 50.26 & 49.10 & 48.96 & 53.75 & 51.29 & 52.65 & 52.87 \\
AVH-Align~\cite{smeu2025circumventing} & \underline{98.94} & \underline{98.53} & \underline{95.94} & \underline{96.54} & 67.92 & 62.36 & 62.24 & 63.38 & \underline{75.92} & 65.87 & 69.70 & 70.94 & 81.44 & 76.52 & 75.59 & 76.76 \\
Human Evaluation & -- & -- & -- & 83.75 & -- & -- & -- & 74.17 & -- & -- & -- & 58.33 & -- & -- & -- & 71.88 \\
\midrule
\midrule
\cellcolor{black!6}\textbf{X-AVDT} (\textbf{Ours}) & \cellcolor{black!6}\textbf{99.10} & \cellcolor{black!6}\textbf{98.89} & \cellcolor{black!6}\textbf{96.54} & \cellcolor{black!6}\textbf{97.05} & \cellcolor{black!6}\underline{90.17} & \cellcolor{black!6}\underline{88.05} & \cellcolor{black!6}\underline{83.11} & \cellcolor{black!6}\underline{84.52} & \cellcolor{black!6}\textbf{97.79} & \cellcolor{black!6}\textbf{97.44} & \cellcolor{black!6}\textbf{97.69} & \cellcolor{black!6}\textbf{97.91} & \cellcolor{black!6}\textbf{95.29} & \cellcolor{black!6}\textbf{94.03} & \cellcolor{black!6}\textbf{91.15} & \cellcolor{black!6}\textbf{91.98} \\
\bottomrule
\end{tabular}
}
\vspace{-0.1cm}
\caption{\textbf{Quantitative comparison on the MMDF dataset.} Detectors are evaluated using the official pretrained checkpoint (first panel) and after retrained on the MMDF training set (Hallo2~\cite{cui2024hallo2}, LivePortrait~\cite{guo2024liveportrait}, and FaceAdapter~\cite{han2024face}) (second panel). Best in \textbf{bold}; second-best \underline{underlined}. \textsuperscript{*}Note: FACTOR is a zero-shot method, while AVAD and AVH-Align are unsupervised methods.}
\label{tab:4_mmdf_comp}
\end{table*}

%% file: table/5_comparison_benchmark.tex
\begin{table}[ht]
\centering
\normalsize
\setlength{\tabcolsep}{2.5pt}
\renewcommand{\arraystretch}{1.1}
\begin{adjustbox}{max width=\columnwidth}
\begin{tabular}{@{}>{\raggedright\arraybackslash}m{2.8cm}
  *4{>{\centering\arraybackslash}m{1.2cm}}
  *4{>{\centering\arraybackslash}m{1.2cm}}
}
\toprule
& \multicolumn{4}{c}{FakeAVCeleb~\cite{khalid2021fakeavceleb}} & \multicolumn{4}{c}{FaceForensics++~\cite{rossler2019faceforensics++}} \\
\cmidrule(lr){2-5}\cmidrule(lr){6-9}
Model & AUROC & AP & Acc@EER & \hspace{5pt}Acc & AUROC & AP & Acc@EER & \hspace{5pt}Acc \\
\midrule
\midrule
\multicolumn{9}{l}{\textit{Official pretrained}} \\
\midrule
LipForensics \cite{haliassos2021lips} & \underline{98.40} & \underline{98.37} & \underline{95.00} & \underline{95.80} & 98.14\textsuperscript{\dag} & 97.93\textsuperscript{\dag} & 97.00\textsuperscript{\dag} & 98.50\textsuperscript{\dag} \\
RealForensics \cite{haliassos2022leveraging} & 95.80 & 96.20 & 85.99 & 85.46 & 99.47\textsuperscript{\dag} & 99.42\textsuperscript{\dag} & 95.49\textsuperscript{\dag} & 95.22\textsuperscript{\dag} \\
AVAD \cite{feng2023self} & 74.96 & 75.88 & 66.00 & 66.79 & 55.38 & 53.92 & 51.50 & 52.16 \\
FACTOR \cite{reiss2023detecting} & 88.44 & 88.22 & 76.00 & 80.00 & 76.33 & 76.08 & 68.50 & 71.00 \\
LipFD \cite{liu2024lips} & 73.17 & 64.39 & 66.36 & 67.16 & 52.84\textsuperscript{\dag} & 51.01\textsuperscript{\dag} & 50.96\textsuperscript{\dag} & 59.31\textsuperscript{\dag} \\
AVH-Align \cite{smeu2025circumventing} & 93.52 & 93.48 & 83.00 & 83.90 & 37.07 & 40.43 & 40.50 & 41.62 \\
\midrule
\multicolumn{9}{l}{\textit{MMDF-trained (cross-dataset evaluation)}} \\
\midrule
RealForensics \cite{haliassos2022leveraging} & 83.67 & 85.56 & 71.42 & 71.42 & \underline{88.85} & \underline{87.65} & \underline{79.39} & \underline{78.54} \\
LipFD \cite{liu2024lips} & 53.59 & 51.09 & 52.37 & 53.67 & 52.92 & 50.21 & 52.08 & 51.89 \\
AVH-Align \cite{smeu2025circumventing} & 54.20 & 53.74 & 52.00 & 52.80 & 36.66 & 40.03 & 37.00 & 39.64 \\
Human Evaluation & -- & -- & -- & 78.75 & -- & -- & -- & 71.25 \\
\midrule
\midrule
\cellcolor{black!6}\textbf{X-AVDT} (\textbf{Ours}) &  \cellcolor{black!6}\textbf{99.69} & \cellcolor{black!6}\textbf{99.74} & \cellcolor{black!6}\textbf{97.85} & \cellcolor{black!6}\textbf{98.65} & \cellcolor{black!6}\textbf{89.55} & \cellcolor{black!6}\textbf{89.17} & \cellcolor{black!6}\textbf{87.55} & \cellcolor{black!6}\textbf{89.77} \\
\bottomrule
\end{tabular}
\end{adjustbox}
\vspace{-0.2cm}
\caption{\textbf{Quantitative comparison on the benchmark dataset.} Detectors are trained on the MMDF training set and evaluated on FakeAVCeleb and FaceForensics++, respectively. \textsuperscript{\dag}Indicates that the corresponding benchmark was used during the method’s original training (train–test overlap).}
\vspace{-0.3cm}
\label{tab:5_ben_comp}
\end{table}

%% file: table/6_ablation_attn.tex
\begin{table*}[t]
\centering
\footnotesize
\setlength{\tabcolsep}{5pt}
\vspace{-0.6cm}
\renewcommand{\arraystretch}{1.05}
\begin{tabular}{l ccc ccc ccc}
\toprule
& \multicolumn{3}{c}{Cross-Attention}
& \multicolumn{3}{c}{Temporal-Attention}
& \multicolumn{3}{c}{Spatial-Attention} \\
\cmidrule(lr){2-4}\cmidrule(lr){5-7}\cmidrule(lr){8-10}
\multicolumn{1}{c}{Timestep $t$}
& AUROC & AP & Acc@EER
& AUROC & AP & Acc@EER
& AUROC & AP & Acc@EER \\
\midrule
\rowcolor{black!6}
$t=24$  & \bfseries 91.56 & \underline{86.90} & \bfseries 81.51 & 83.92 & \bfseries 87.08 & 75.01 & 64.57 & 68.00 & 61.82 \\
$t=249$ & 81.30 & 69.52 & 77.13 & 68.25 & 71.94 & 64.10 & 57.42 & 52.00 & 56.04 \\
$t=499$ & 68.11 & 57.97 & 67.44 & 66.29 & 66.29 & 59.30 & 52.38 & 53.14 & 50.02 \\
\bottomrule
\end{tabular}
\vspace{-0.2cm}
\caption{\textbf{Ablation results for different attention features and diffusion timesteps.} As the diffusion timestep $t$ grows, the latent becomes noisier and conditioning weakens. Results at $t\in\{24,249,499\}$ show that cross-attention performs best, and performance degrades as $t$ increases.}
\vspace{-0.2cm}
\label{tab:6_ablation_attn_feat}
\end{table*}

%% file: figs/2_ablation_attn.tex
\begin{figure}[t]
 \includegraphics[width=\columnwidth]{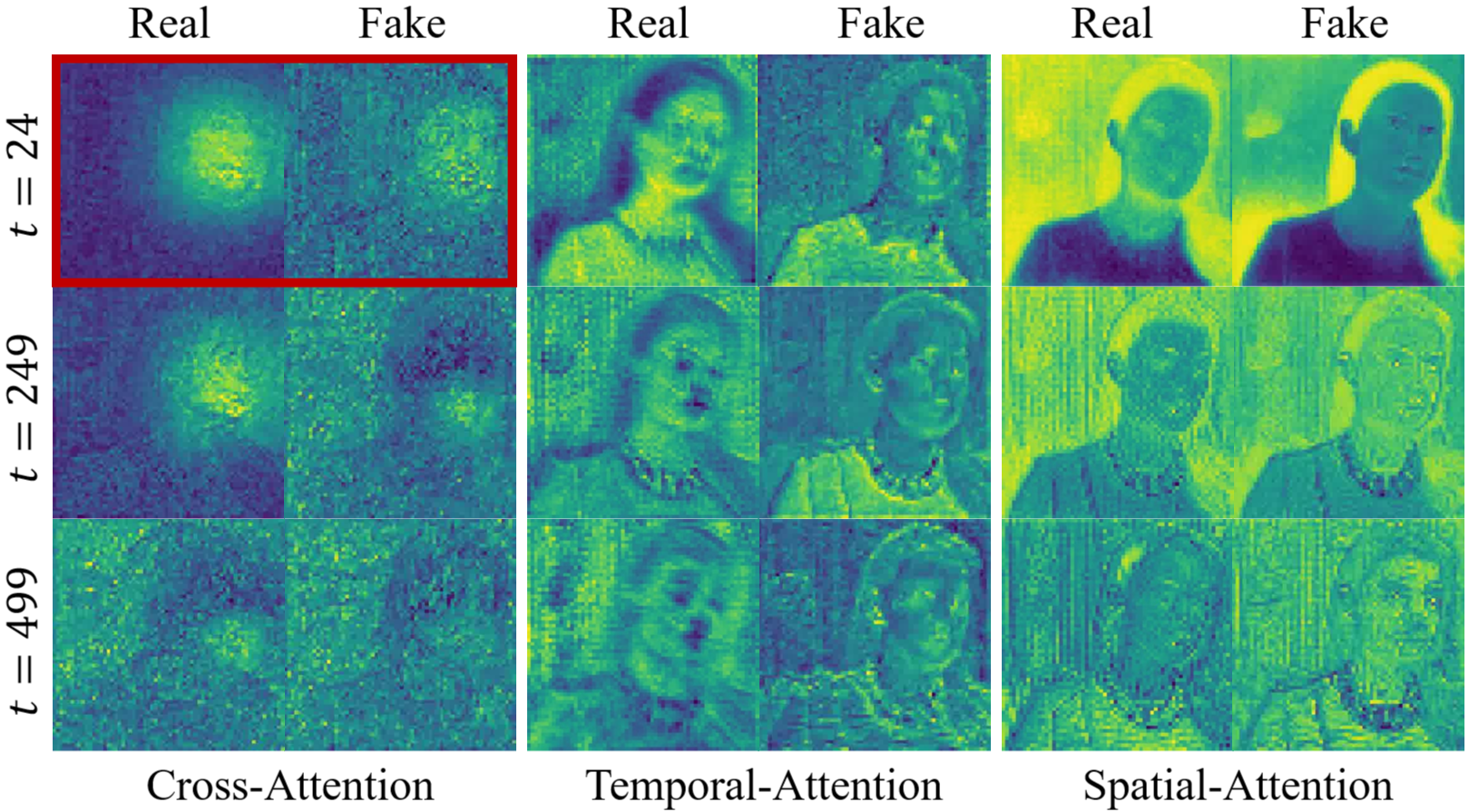}
 \vspace{-0.6cm}
 \caption{\textbf{Comparison of attention features across diffusion timesteps.} Red box denote the configuration used in our method.}
 \vspace{-0.2cm}
 \label{fig:abl_attn}
\end{figure}

%% file: figs/5_perturbation.tex
\begin{figure*}
 \vspace{-0.8cm}
 \begin{center}
 \includegraphics[width=0.9\linewidth]{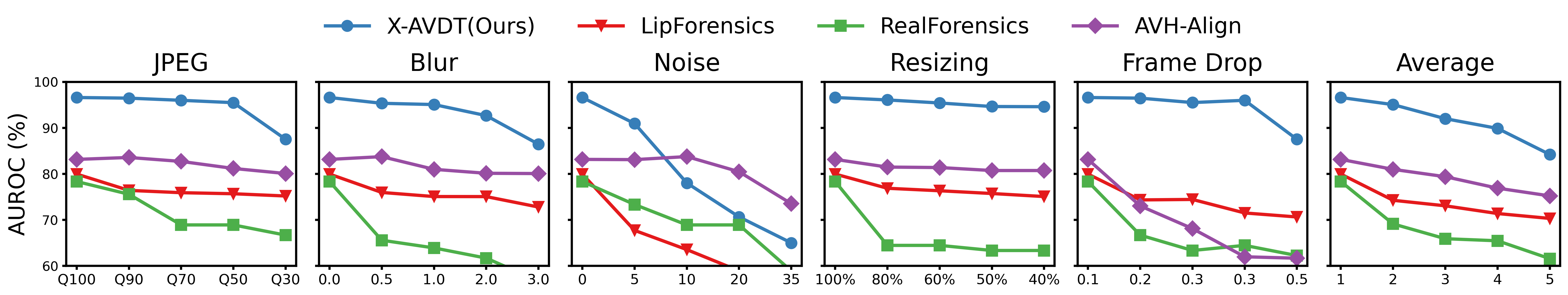}
 \end{center}
 \vspace{-0.6cm}
 \caption{\textbf{Robustness against unseen corruptions.} AUROC (\%) across five severity levels. Per corruption scales are shown on the $x$–axes. Average denotes the mean AUROC across all corruptions at each severity level.}
 \vspace{-0.1cm}
 \label{fig:5_abl_perturbation}
\end{figure*}

%% file: figs/4_grad_cam.tex
\begin{figure}
 \includegraphics[width=\linewidth]{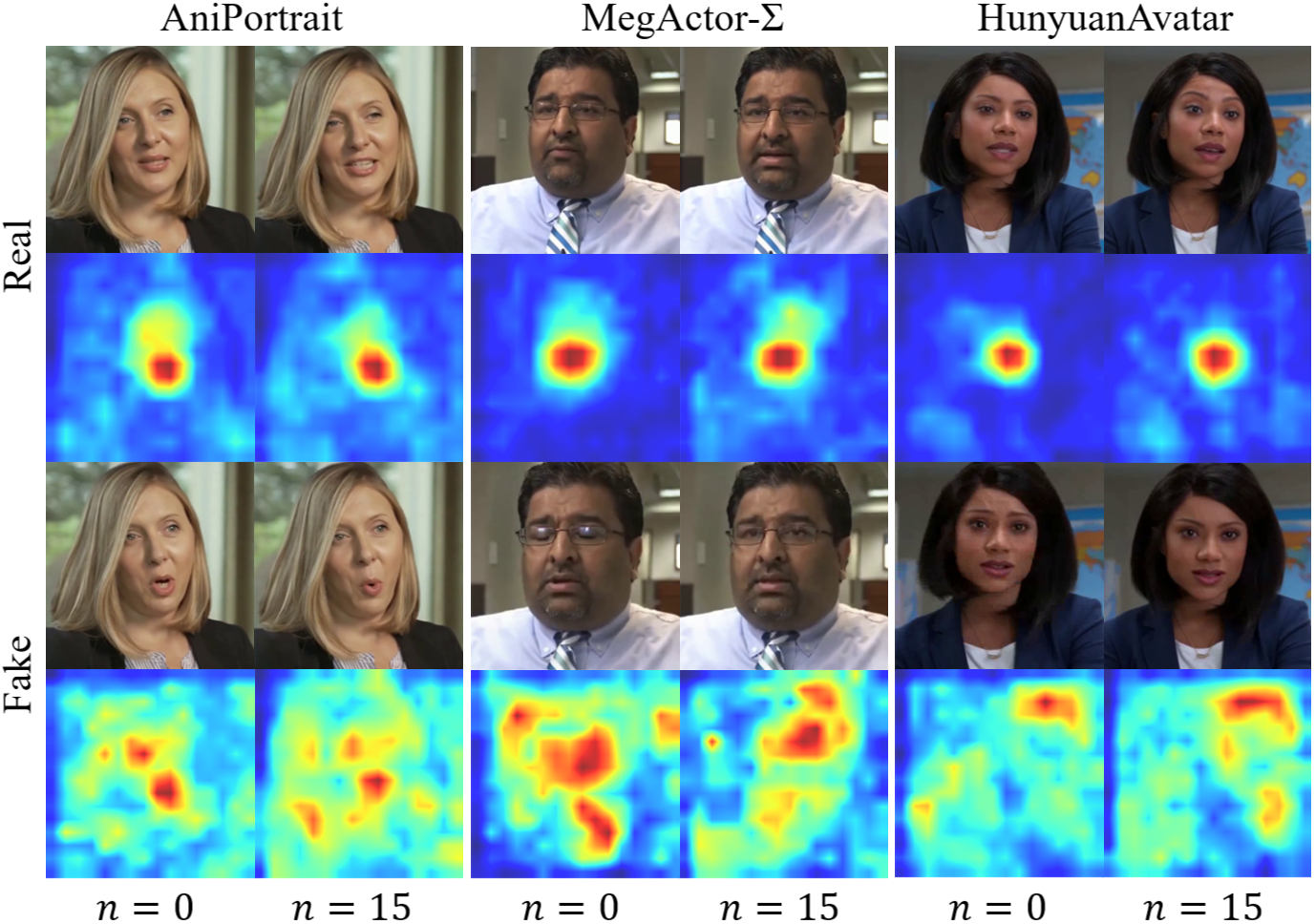}
 \vspace{-0.6cm}
 \caption{\textbf{Grad-CAM visualizations.} Red activation indicates regions where our model focuses most (i.e., pixels that make a strong positive contribution to the predicted class), while cooler colors denote weak or no contribution.}
 \vspace{-0.1cm}
 \label{fig:grad_cam}
\end{figure}

%% file: table/7_ablation_input_loss.tex
\begin{table}[t]
\centering
\footnotesize
\resizebox{\linewidth}{!}{
\begin{tabular}{lccc}
\toprule
Method & AUROC & AP & Acc@EER \\
\midrule
\multicolumn{4}{l}{(a) Ablation on Input Representations}\\
\midrule
\textit{w/o} AV Cross-Attn ($\boldsymbol{\psi}$) & 88.22 & 87.25 & 83.70 \\
\textit{w/o} Video Composite ($\boldsymbol{\phi}$) & 90.21 & 90.57 & 84.32 \\
\textit{w/o} Residual ($\lvert x-D(\hat z_0)\rvert$) & 93.82 & 92.25 & 89.00 \\
\midrule
\multicolumn{4}{l}{(b) Ablation on Loss Design}\\
\midrule
\textit{w/o} $\mathcal{L}_{\text{tri}}$ & 92.64 & 92.26 & 86.32 \\
\rowcolor{black!6} 
\textbf{X-AVDT} (\textbf{full}) & \textbf{95.29} & \textbf{94.03} & \textbf{91.15} \\
\bottomrule
\end{tabular}
}
\vspace{-0.2cm}
\caption{\textbf{Ablation on input representations and loss design.} (a) removing any of the input representation degrades performance,} (b) adding the triplet term improves the results across all metrics.
\vspace{-0.3cm}
\label{tab:7_abl_input_loss}
\end{table}

%% file: sec/7_conclusion.tex
\section{Conclusion}
\vspace{-0.1cm}
We introduce X-AVDT, a simple, robust, and generalizable detector that probes internal audio–visual signals in pretrained diffusion models via DDIM inversion. Our approach fuses two complementary representations: a video composite $\boldsymbol{\phi}$ that surfaces reconstruction-based discrepancies and an AV cross-attention feature $\boldsymbol{\psi}$ that encodes speech–motion synchrony. Empirically, we demonstrate that probing intermediate diffusion features during inversion yields more discriminative and better calibrated signals than relying solely on end point reconstructions. Evaluated on MMDF and external benchmarks, X-AVDT delivers consistent improvements over prior methods and transfers well to unseen generators, achieving superior performance on standard datasets and under perturbed image conditions. We also present MMDF, an audio–visual deepfake benchmark curated for cross-generator generalization and robustness studies. We hope MMDF serves as a strong benchmark suite for advancing detector generalization and real-world robustness, while X-AVDT provides a solid baseline and diagnostic probe for future work on audio–visual, generator-internal cues.

\noindent\textbf{Limitations and Future Work} Despite strong accuracy and cross-dataset robustness, our method incurs a high computational cost, reflecting the inherent expense of inversion. For a 16 frame clip, a full DDIM inversion and reconstruction process with a 40 timestep schedule requires approximately one minute end-to-end. Current detectors also remain imperfect in non-speech segments or multi-speaker scenes, because the approach relies on speech-driven features. A few promising directions include supplementing our representation with unimodal back-off strategies for weak or missing speech and with speech-agnostic correspondence cues beyond phonetics, and pursuing lightweight inversion via distillation.

%% file: sec/X_suppl.tex
\clearpage
\appendix
\setcounter{page}{1}
\maketitlesupplementary 

\input{sup_sec/0_index}
\input{sup_sec/A_experimental_details}

\input{sup_sec/B_additional_experiments}
\input{sup_sec/C_ablation_study}
\input{sup_sec/D_Analysis}
\input{sup_sec/E_dataset}

%% file: sup_sec/0_index.tex
\noindent In this supplementary material, we provide expanded details on the proposed model and the data, an extended ablation study, a class separability analysis, and additional visualizations:
\begin{itemize}
\item In Section~\ref{sup:a}, we present additional technical details of our training setup, including the inversion procedure, attention feature extraction and the model architecture. We also describe how the baselines were trained, and detail the human evaluation.
\item In Section~\ref{sup:b}, we report the results of experiments on broader deepfake benchmarks, and provide comparative analyses with representative audio-visual baselines.
\item In Section~\ref{sup:c}, we (i) report the results of an extended ablation study that compares inversion conditions (audio-driven, text-driven, and withou inversion), (ii) provide detailed results under perturbation attacks, and (iii) analyze attention types and diffusion timesteps.
\item In Section~\ref{sup:d}, we conduct a class separability analysis using Fisher SNR and LDA margin to quantify the discriminability of the learned representations. We also present extended cross-attention robustness analyses, along with attention map visualizations that support these findings.
\item In Section~\ref{sup:e}, we describe how the MMDF training and evaluation data were obtained and present qualitative examples, including our model’s input representations and sample dataset visualizations.
\item In Section~\ref{sup:f}, we discuss the limitations of our system.
\end{itemize}

%% file: sup_sec/A_experimental_details.tex
\vspace{-2mm}
\section{Additional Experimental Details}
\label{sup:a}
\vspace{-2mm}
\subsection{Implementation Details of X-AVDT}
\vspace{-1mm}
\subsubsection{Input Representation}
The full procedure of input representation extraction is summarized in Algorithm~\ref{alg:input}. We follow Hallo~\cite{xu2024hallo} with a paired ReferenceNet~\cite{hu2024animate} to encode identity features from the source portrait frame. During DDIM inversion, the denoising U-Net reads these features via cross-attention. For the cross-attention feature in our detector, we sample at an early diffusion step, setting $t^\star\!=\!24$ out of a 1000 step schedule during inversion. We adopt Hallo's hierarchical audio-visual cross-attention mechanism to handle regional masking. We compute lip, expression, and pose masks, apply them as element-wise gates to the cross-attention features and then aggregate the gated features with learned weights. We operate clip-wise on non-overlapping 16 frame segments. If the video length is not divisible by 16, we repeat the last frame to pad to the nearest multiple before feature extraction, and concatenate the per-clip outputs along time. This extraction pipeline is applied identically across all datasets for both training and evaluation.

\begin{algorithm}
\small
\DontPrintSemicolon
\caption{Audio-driven inversion \& reconstruction.}
\label{alg:input}
\KwIn{Video $x$, Reference frame $x_{\mathrm{ref}}$, Audio $c$, \\
\quad\quad\quad Masks $\mathcal M=\{\mathcal M_{\mathrm{full}},\mathcal M_{\mathrm{face}},\mathcal M_{\mathrm{lip}}\}$}

\KwOut{Video composite $\boldsymbol{\phi}=[x, D(\hat z_T), D(\hat z_0), r]$,\\
\quad\quad\quad AV cross-attention feature $\boldsymbol{\psi}=\mathrm{CrossAttn}(H(t),\,c)$}

\textbf{1. Encode.}
$z_0 \leftarrow \mathrm{VAE_{\mathrm{enc}}}(x)$,\quad
$e_a \leftarrow \mathrm{Audio_{\mathrm{enc}}}(c)$\;

\textbf{2. Reference pass.}
$\mathrm{RefFeat} \leftarrow \mathrm{ReferenceNet}(x_{\mathrm{ref}})$\;

\textbf{3. DDIM Inversion.} Run the inverse scheduler $z_0 \rightarrow z_T$.\;

    \For{$t \in T$ (fine $\rightarrow$ coarse)}{
      $\bigl(\hat\epsilon_t,\ {\boldsymbol{\psi}}_t\bigr) \leftarrow
        \mathrm{UNetFwd}\!\left(
          z_t,\ e_a,\ \mathcal M;\ \mathrm{RefFeat}\right)$\;
          
      $z_{t+1} \leftarrow \mathrm{DDIMInverseScheduler}(z_t,\hat\epsilon_t)$\;
      
      \If{$t = t^\star$}{
        $\tilde{\boldsymbol{\psi}} \leftarrow \mathrm{HeadProj}(\psi_t)$ 
        
        $\boldsymbol{\psi} \leftarrow \sum_{k\in\{\mathrm{full,face,lip}\}} w_k\,(\tilde{\boldsymbol{\psi}} \odot \mathcal M_k)$ }}

\mbox{\textbf{4. DDIM Reconstruction.} Run the forward scheduler $z_T \rightarrow z_0$.}
    \For{$t \in T$ (coarse $\rightarrow$ fine)}{
      $\hat\epsilon_t \leftarrow \mathrm{UNetFwd}(\tilde z_t,\, e_a,\, \mathcal M;\, \mathrm{RefFeat})$

      $\tilde z_{t-1} \leftarrow \mathrm{DDIMScheduler}(\tilde z_t,\, \hat\epsilon_t)$}

$\hat z_0 \leftarrow \tilde z_0$, \,
$\hat x \leftarrow D(\hat z_0)$, \,
$u \leftarrow D(\hat z_T)$, \,
$r \leftarrow |x-\hat x|$, \,

$\boldsymbol{\phi} \leftarrow [x,\,u,\,\hat x,\,r]$\;

\textbf{Return} $(\boldsymbol{\phi},\ \boldsymbol{\psi})$
\end{algorithm}

\subsubsection{Conditioning}
We use wav2vec 2.0~\cite{baevski2020wav2vec} as the audio feature encoder to condition our videos. To capture rich semantics information across different audio layers, we concatenate the audio embeddings from the last 12 layers of wav2vec 2.0 network. Given the sequential nature of audio, we aggregate a 5-frame local context (t$-$2…t$+$2) for each video frame before projection.

\subsubsection{Training}
To fuse the video composite $\boldsymbol{\phi}$ and AV cross-attention feature $\boldsymbol{\psi}$ during training, we proceed as follows. We concatenate $\mathbf{v}'$ and $\mathbf{a}'$ along the channel dimension and apply a $1{\times}1$ convolution, reducing the channels from $2048$ to $1024$ to obtain $p_i$. We add fixed 2D positional encodings to $p_i$ and apply an 8-head self-attention over the $H\!W$ tokens, with LayerNorm and a residual connection. We feed the self-attention outputs into three 3D ResNeXt~\cite{xie2017aggregated} layers, followed by global average pooling, which yields $g_i\in\mathbb{R}^{1024}$. We train for 2 epochs by default, as our inputs are structured internal representations extracted from a pretrained diffusion model, enabling faster convergence than raw RGB. Table~\ref{tab:epoch} reports an ablation result showing that performance quickly converges after a few epochs.

\begin{table}[th]
\centering
\renewcommand{\thetable}{A.1.3}
\vspace{-10mm}
\scalebox{0.9}{
\begin{tabular}{lccccc}
    \toprule
    Epochs & 1 &  2 (Ours) & 5 & 10 & 20   \\ \midrule
    AUROC  & 93.24 & \textbf{95.29} & 95.01 & 95.19 & 95.13 
    \\ \bottomrule
\end{tabular}}
\caption{\textbf{Effect of training epochs on X-AVDT.}}
\vspace{-2mm}
\label{tab:epoch}
\end{table}

\subsection{Details of Baseline Detectors}
\subsubsection{\texorpdfstring{LipForensics~\cite{haliassos2021lips}}{LipForensics}}
\vspace{-1mm}
LipForensics is a video-only deepfake detector that operates on mouth crops, targeting the lip region and modeling temporal inconsistencies in mouth movements to identify manipulation-specific irregularities. We evaluated LipForensics using the official pretrained model that has been trained on FaceForensics++. In addition, we did not retrain it because the training code is not available.

\subsubsection{\texorpdfstring{RealForensics~\cite{haliassos2022leveraging}}{RealForensics}}
\vspace{-1mm}
RealForensics uses audio-visual pretraining, in which audio and visuals exclusively from real samples are used to learn representations that help a classifier discriminate between real and fake videos. We evaluated RealForensics using the official pretrained model that has been trained on FaceForensics++, and we also retrained it on MMDF using the same hyperparameters.

\subsubsection{\texorpdfstring{AVAD~\cite{feng2023self}}{AVAD}}
\vspace{-1mm}
AVAD first pretrains an audio–visual synchronization model following  Chen \textit{et al.}~\cite{chen2021audio} to learn temporal alignment between speech and mouth motion. They then use the inferred features to train an anomaly detector, producing a fully unsupervised multi-modal deepfake detector. As an unsupervised method, AVAD is not trained with labels or fake examples. We evaluated AVAD using the official pretrained model that was trained on LRS~\cite{son2017lip} and did not retrain it because the training code is not available.

\subsubsection{\texorpdfstring{FACTOR~\cite{reiss2023detecting}}{FACTOR}}
\vspace{-1mm}
FACTOR is a training-free deepfake detector that frames detection as fact checking. It uses audio-visual encoders to extract modality-specific features and computes a truth score (cosine similarity) that quantifies the consistency between observed audio-visual evidence and an asserted attribute. FACTOR operates in a zero-shot, label-free setting and does not use fake examples. We evaluated FACTOR using the official implementation and pretrained feature extractors.

\subsubsection{\texorpdfstring{LipFD~\cite{liu2024lips}}{LipFD}}
\vspace{-1mm}
LipFD targets lip-syncing forgeries by enforcing audio-visual temporal consistency between lip movements and audio signals. The method operates on mouth crops and combines a global video branch with a lip-region branch in a dual-head design. We evaluated LipFD using the model pretrained on Lip Reading Sentences 3 (LRS3)~\cite{afouras2018lrs3}, FaceForensics++, and the Deepfake Detection Challenge Dataset (DFDC)~\cite{dolhansky2020deepfake}. For retraining on MMDF, we trained LipFD for 25 epochs and otherwise keep the original hyperparameters.


\subsubsection{\texorpdfstring{AVH-Align~\cite{smeu2025circumventing}}{AVH-Align}}
\vspace{-1mm}
AVH-Align addresses dataset shortcuts such as leading silence by training only on real data and learning a frame-level audio-video alignment score from AV-HuBERT features~\cite{shi2022learning}. The training is self-supervised and label-free on real pairs, with no fake examples used. We evaluated AVH-Align using the official pretrained model that has been trained on FakeAVCeleb and AV-Deepfake1M. In addition, we retrained it on MMDF using the same hyperparameters.
\vspace{-1mm}

\begin{table}[t]
    \renewcommand{\thetable}{B.1}
    \centering
    \vspace{-10mm}
    \scalebox{0.7}{
        \begin{tabular}{@{} *3{c} c *3{c} @{}}
            \toprule
            \multicolumn{3}{c}{FakeAVCeleb~\cite{khalid2021fakeavceleb}} &  & \multicolumn{3}{c}{FaceForensics++~\cite{rossler2019faceforensics++}} \\
            \cmidrule(lr){1-3}\cmidrule(lr){5-7}
            FSGAN & FaceSwap & Wav2Lip &  & Deepfakes  & FaceSwap & Face2Face  \\
            \midrule
            99.73 & 99.79 & 99.92 &  & 99.62 & 99.24 & 99.63 \\
            \bottomrule
        \end{tabular}
    }
    \caption{\textbf{In-domain AUROC on the benchmark dataset.} Cross-manipulation generalization is evaluated by training on two manipulation methods and testing on the remaining one (e.g., train on the first two columns and test on the third columns).}
    \vspace{-1mm}
    \label{tab:5_ben_comp_indomain}
\end{table}

\vspace{-5mm}
\subsection{Human Evaluation}
\vspace{-1mm}
We conducted a human evaluation study to assess deepfake detection accuracy and to quantify the realism of manipulated videos in MMDF, comparing results against FaceForensics++ and FakeAVCeleb. For each clip, participants responded to two following questions: (i) \emph{“Is the video real or fake?"} (binary choice), and (ii) \emph{"What did you focus on when deciding whether the video was real or fake?"}. For question (ii), 80\% of comments cited audio-visual synchronization as the primary cue, while the remainder pointed to background artifacts, expression dynamics, and intraoral details, etc. We used 80 videos (60 from the MMDF dataset, 10 from the FakeAVCeleb, and 10 from FaceForensics++), with 24 participants providing responses. We aggregated answers to compute Human Evaluation (HE) accuracy and Human False Acceptance Rate (HFAR), with both metrics derive from the same study.

%% file: sup_sec/B_additional_experiments.tex
\vspace{-2mm}
\section{Additional Experiments}
\label{sup:b}
\vspace{-1mm}
\subsection{In-domain Evaluation}
\vspace{-1mm}
Table~\ref{tab:5_ben_comp_indomain} summarizes the in-domain performance of X-AVDT on each benchmark dataset (FakeAVCeleb~\cite{khalid2021fakeavceleb} and FaceForensics++~\cite{rossler2019faceforensics++}), where the model is trained and tested on the same dataset ($\sim$ \textit{Official-pretrained}). X-AVDT achieved high AUROC across manipulation types (higher than the scores obtained by the prior methods presented in Table~\ref{tab:5_ben_comp}). Note that Table~\ref{tab:5_ben_comp} reports the results for cross-dataset robustness (MMDF $\rightarrow$ benchmark); while the performances of many \textit{MMDF-trained} baselines dropped due to domain mismatch, X-AVDT remained strong.

\begin{table*}[h]
\centering
\setcounter{table}{0}
\renewcommand{\thetable}{B.2.1}
\vspace{-10mm}
\setlength{\tabcolsep}{2pt}  
\resizebox{\linewidth}{!}{
\begin{tabular}{@{}>{\raggedright\arraybackslash}m{3.1cm}
  *3{>{\centering\arraybackslash}m{1.5cm}}
  *3{>{\centering\arraybackslash}m{1.5cm}}
  *3{>{\centering\arraybackslash}m{1.5cm}}
  *3{>{\centering\arraybackslash}m{1.5cm}}
}
\toprule
& \multicolumn{3}{c}{AniPortrait~\cite{wei2024aniportrait}} & \multicolumn{3}{c}{MegActor-$\Sigma$~\cite{yang2025megactor}} & \multicolumn{3}{c}{HunyuanAvatar~\cite{chen2025hunyuanvideo}} & \multicolumn{3}{c}{Average} \\
\cmidrule(lr){2-4}\cmidrule(lr){5-7}\cmidrule(lr){8-10}\cmidrule(lr){11-13}
Model & AUROC & AP & Acc@EER & AUROC & AP & Acc@EER & AUROC & AP & Acc@EER & AUROC & AP & Acc@EER \\
\midrule
SpeechForensics~\cite{liang2024speechforensics} & 99.99 & 99.99 & 99.88 & 98.69 & 98.82 & 94.46 & 92.12 & 91.98 & 82.90 & 96.93 & 96.93 & 92.41 \\
\midrule
\cellcolor{black!6}\textbf{X-AVDT} (\textbf{Ours}) & \cellcolor{black!6}\textbf{99.10} & \cellcolor{black!6}\textbf{98.89} & \cellcolor{black!6}\textbf{96.54} & \cellcolor{black!6}\textbf{90.17} & \cellcolor{black!6}\textbf{88.05} & \cellcolor{black!6}\textbf{83.11} & \cellcolor{black!6}\textbf{97.79} & \cellcolor{black!6}\textbf{97.44} & \cellcolor{black!6}\textbf{97.69} & \cellcolor{black!6}\textbf{95.29} & \cellcolor{black!6}\textbf{94.03} & \cellcolor{black!6}\textbf{91.15} \\
\bottomrule
\end{tabular}
}
\caption{\textbf{Additional quantitative comparison results on the MMDF dataset.}}
\label{tab:b_add_com}
\end{table*}

\begin{table}[t]
    \centering
    \renewcommand{\thetable}{B.2.2}
    \scalebox{0.8}{
        \begin{tabular}{lcccc}
            \toprule
            \textit{Dataset} &  AUROC & AP & Acc@EER & Acc \\ \midrule
            DeepSpeak v1.0~\cite{barrington2024deepspeak} & 94.29 & 95.06 & 95.39 & 94.94 \\
            KoDF~\cite{kwon2021kodf} & 93.07 & 92.88 & 91.13 & 91.87 \\  
            Deepfake-Eval2024~\cite{chandra2025deepfake} & 75.02 & 72.73 & 71.68 & 71.36 \\
            \bottomrule
        \end{tabular}
    }
    \caption{\textbf{Additional quantitative results for X-AVDT.}}
    \label{tab:add_dataset}
\end{table}

\subsection{Additional Experiments}
In addition to Table~\ref{tab:4_mmdf_comp} in the main paper, we report additional quantitative results to further evaluate the generalization of SpeechForensics~\cite{liang2024speechforensics}. Table~\ref{tab:b_add_com} compares X-AVDT with SpeechForensics using representative audio-visual baselines on the MMDF dataset. Although X-AVDT performed worse than SpeechForensics on AniPortrait and MegActor-$\Sigma$, it yielded a clear improvement on HunyuanAvatar, where SpeechForensics attained comparatively lower scores. Overall, the results indicate complementary strengths of the two methods across generators and suggest that generator-internal audio-visual consistency cues can be particularly helpful in challenging settings such as HunyuanAvatar, whose high-fidelity, temporally coherent, audio-driven synthesis can suppress overt artifact-based cues. Table~\ref{tab:add_dataset} summarizes results on additional in-the-wild deepfake benchmarks, such as DeepSpeak v1.0~\cite{barrington2024deepspeak}, KoDF~\cite{kwon2021kodf}, and Deepfake-Eval2024~\cite{NIST2025DeepfakeEval}. X-AVDT maintained high performance on DeepSpeak v1.0 and KoDF, but its performance dropped on Deepfake-Eval2024, due to a challenging domain shift in content and compression conditions. Nevertheless, the method remained well above chance across all three datasets, indicating non-trivial generalization in settings with MMDF.

%% file: sup_sec/C_ablation_study.tex
\section{Additional Ablation Study}
\label{sup:c}

\begin{table*}[h]
\centering
\setcounter{table}{0}
\renewcommand{\thetable}{C.\arabic{table}}
\setlength{\tabcolsep}{2pt}  
\resizebox{\linewidth}{!}{
\begin{tabular}{@{}>{\raggedright\arraybackslash}m{3.1cm}
  *3{>{\centering\arraybackslash}m{1.5cm}}
  *3{>{\centering\arraybackslash}m{1.5cm}}
  *3{>{\centering\arraybackslash}m{1.5cm}}
  *3{>{\centering\arraybackslash}m{1.5cm}}
}
\toprule
& \multicolumn{3}{c}{AniPortrait~\cite{wei2024aniportrait}} & \multicolumn{3}{c}{MegActor-$\Sigma$~\cite{yang2025megactor}} & \multicolumn{3}{c}{HunyuanAvatar~\cite{chen2025hunyuanvideo}} & \multicolumn{3}{c}{Average} \\
\cmidrule(lr){2-4}\cmidrule(lr){5-7}\cmidrule(lr){8-10}\cmidrule(lr){11-13}
Method & AUROC & AP & Acc@EER & AUROC & AP & Acc@EER & AUROC & AP & Acc@EER & AUROC & AP & Acc@EER \\
\midrule
\textit{w/o} Inversion & 73.55 & 67.56 & 69.95 & 63.18 & 63.69 & 59.81 & 41.81 & 46.36 & 43.55 & 62.22 & 61.07 & 56.69 \\
Text-driven & 90.22 & 90.85 & 81.56 & 76.15 & 75.25 & 69.91 & 49.01 & 49.00 & 50.51 & 74.48 & 76.94 & 65.75 \\
\midrule
\cellcolor{black!6}\textbf{Audio-driven} (\textbf{Ours}) & \cellcolor{black!6}\textbf{96.55} & \cellcolor{black!6}\textbf{98.83} & \cellcolor{black!6}\textbf{90.20} & \cellcolor{black!6}\textbf{89.87} & \cellcolor{black!6}\textbf{89.25} & \cellcolor{black!6}\textbf{83.86} & \cellcolor{black!6}\textbf{97.71} & \cellcolor{black!6}\textbf{97.04} & \cellcolor{black!6}\textbf{90.94} & \cellcolor{black!6}\textbf{94.71} & \cellcolor{black!6}\textbf{95.04} & \cellcolor{black!6}\textbf{88.33} \\
\bottomrule
\end{tabular}
}
\caption{\textbf{Ablation on video composite $\boldsymbol{\phi}$}. When training, we fix the backbone and vary only the conditioning signal used during inversion. Audio-driven setting ranked first across datasets, while removing inversion cues yielded the weakest composite.} 
\label{tab:b_abl_condition}
\end{table*}

\begin{figure*}
\setcounter{figure}{0}
\renewcommand{\thefigure}{C.2.1}
 \begin{center}
 \includegraphics[width=0.9\linewidth]{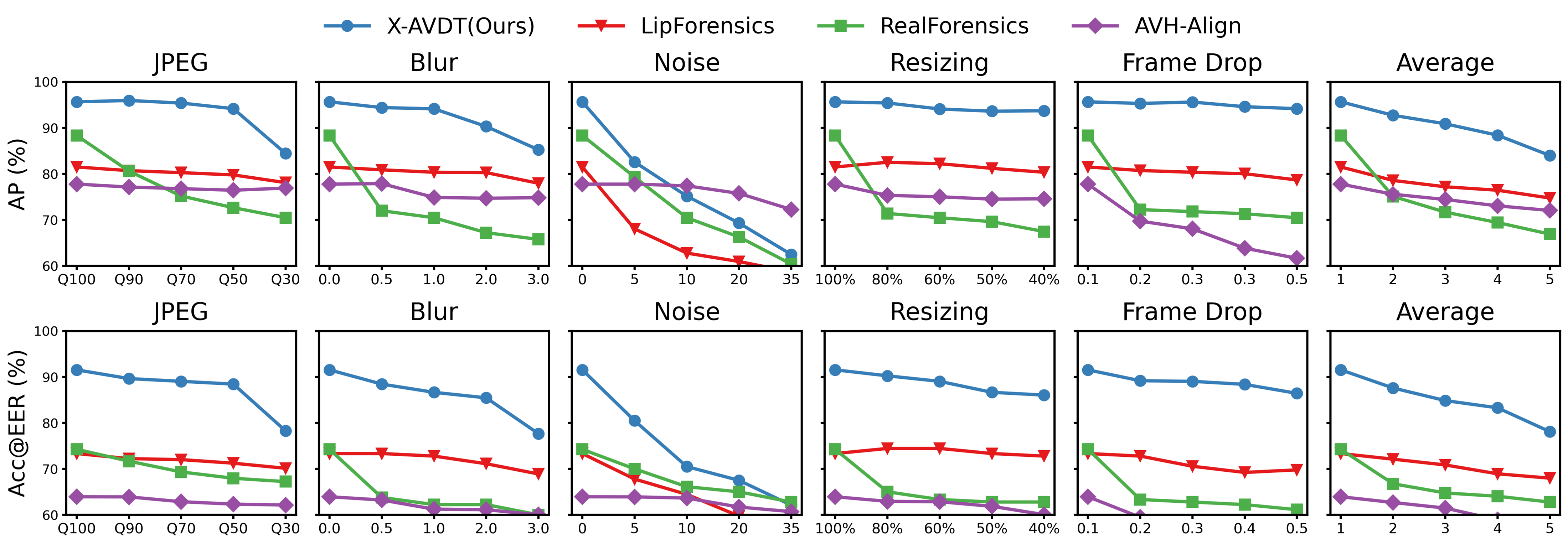}
 \end{center}
 \caption{\textbf{Robustness to unseen corruptions.} AP (\%) is shown in the first row and Acc@EER (\%) is shown in the second row across five severity levels. Per corruption scales are shown on the $x$–axes. Average denotes the mean AP in the top row and Acc@EER in the bottom row, respectively, across all corruptions at each severity level.}
 \label{fig:b2_abl_perturbation}
\end{figure*}

\begin{table}[h]
\centering
\setcounter{table}{0}
\renewcommand{\thetable}{C.2.2}
\setlength{\tabcolsep}{2pt}  
\resizebox{\linewidth}{!}{
\begin{tabular}{@{}>{\raggedright\arraybackslash}m{3.1cm}
  *3{>{\centering\arraybackslash}m{1.5cm}}
  *3{>{\centering\arraybackslash}m{1.5cm}}
}
\toprule
& \multicolumn{3}{c}{Audio Desynchronizaiton} & \multicolumn{3}{c}{Audio Codec Artifacts} \\
\cmidrule(lr){2-4}\cmidrule(lr){5-7}
Metric & -0.5 sec & 0 & +0.5 sec & 0 & 8k & 32k \\
\midrule
AUROC & 90.90 & 93.70 & 91.31 & 93.70 & 91.80 & 90.17 \\
AP & 91.10 & 94.30 & 91.74 & 94.30 & 88.80 & 86.51 \\
Acc@EER & 83.40 & 86.40 & 83.97 & 86.40 & 85.90 & 81.56 \\
\bottomrule
\end{tabular}
}
\caption{\textbf{Robustness to unseen audio perturbations.} Performance under audio desynchronization (temporal offsets) and audio codec artifacts (low-bitrate re-encoding) at varying severity levels.}
\label{tab:abl_audio_perturb}
\end{table}

\subsection{Inversion Condition}
In Table~\ref{tab:b_abl_condition}, we compare three input settings for the video composite $\boldsymbol{\phi}$. The settings are: text-driven conditioning; original-frame only, which uses only the original RGB frames without the decoded latent DDIM noise map $D(\hat z_T)$, the reconstruction $D(\hat z_0)$, the residual $r=\lvert x-D(\hat z_0)\rvert$, or attention features; and audio-driven conditioning (Ours). For text-driven conditioning, we use the BLIP-2, OPT-2.7b~\cite{li2023blip} model to caption frames before inversion. The text-conditioned inversion is based on Stable Diffusion 1.5~\cite{rombach2022high}, the same backbone as Hallo. This ablation represents the core designing principle of X-AVDT leveraging internal features of large generative models, specifically audio-visual cross-attention features for deepfake detection. The audio-driven setting consistently yielded the strongest results, validating that audio conditioned cross-attention offers richer, temporally aligned cues than text conditioning. Such alignment is particularly important for facial-editing videos that hinge on subtle mouth and expression edits.

\vspace{-4mm}
\subsection{Robustness of Perturbation Attack}
\vspace{-2mm}
We conducted an additional ablation study to evaluate the robustness of X-AVDT exploiting audio–visual alignment signals against diverse image perturbations. We assessed performance under five scenarios, with severity 0 denoting the unmodified original. All experiments were run on a subset of MMDF. The baselines (LipForensics~\cite{haliassos2021lips}, RealForensics~\cite{haliassos2022leveraging}, and AVH-Align~\cite{smeu2025circumventing}) were evaluated with their official pretrained checkpoints.
\begin{itemize}
\item \textbf{JPEG Compression}: Lossy re-encoding is applied with quality levels of 90, 70, 50, and 30. Lower quality yields stronger high-frequency suppression.
\item \textbf{Blur}: Gaussian blur with radius of 0.5, 1.0, 2.0, and 3.0, modeling defocus and motion smoothing.
\item \textbf{Noise}: Additive Gaussian noise with standard deviation (pixel scale) of 5, 10, 20, and 35.
\item \textbf{Resizing}: Downscale and then upscale using bilinear interpolation with percentage of  75\%, 60\%, 50\%, and 40\%. The frame is reduced and then upsampled back to the original size, simulating resolution loss.
\item \textbf{Frame Drop}: Randomly remove frames with probabilities of 0.05, 0.10, 0.20, and 0.30, creating temporal discontinuities. We do not duplicate frames in this setting.
\end{itemize}
These experiments were conducted with the same dataset used in Table~\ref{tab:4_mmdf_comp} and Table~\ref{tab:5_ben_comp}. The quantitative results for AP (\%) and Acc@EER (\%) are presented in Figure~\ref{fig:b2_abl_perturbation}. As shown in the results, X-AVDT caused only minor performance degradation across various perturbation methods, while consistently surpassed competing baselines~\cite{haliassos2021lips,haliassos2022leveraging,smeu2025circumventing}, demonstrating strong robustness.

\noindent{\textbf{Audio Perturbation Attack.}}
We evaluates the robustness of X-AVDT under two audio perturbations: Audio Desynchronization and Audio Codec Artifacts. For desynchronization, we apply a temporal offset $\tau\in\{-0.5,+0.5\}$ seconds, where a positive offset delays speech by prefixing $|\tau|$ seconds of silence, while a negative offset advances audio by trimming the first $|\tau|$ seconds. For codec artifacts, we re-encode audio with a bitrate cap $b\in\{8,32\}$ kbps to introduce compression distortions. As shown in Table~\ref{tab:abl_audio_perturb}, compared to the clean setting ($\tau=0$, original audio), desynchronization induced only modest performance drops (within 2.4-3.2 points across metrics) at $\tau=\pm0.5$s. Similarly, compared to no re-encoding, codec artifacts caused limited degradation (within 0.5-7.8 points across metrics) under $b\in\{8,32\}$ kbps. These results suggest that the learned audio-visual consistency cues remain stable under temporal misalignment and compression-induced distortions.

\begin{figure*}
\setcounter{figure}{0}
\renewcommand{\thefigure}{C.3}
 \begin{center}
 \includegraphics[width=0.9\linewidth]{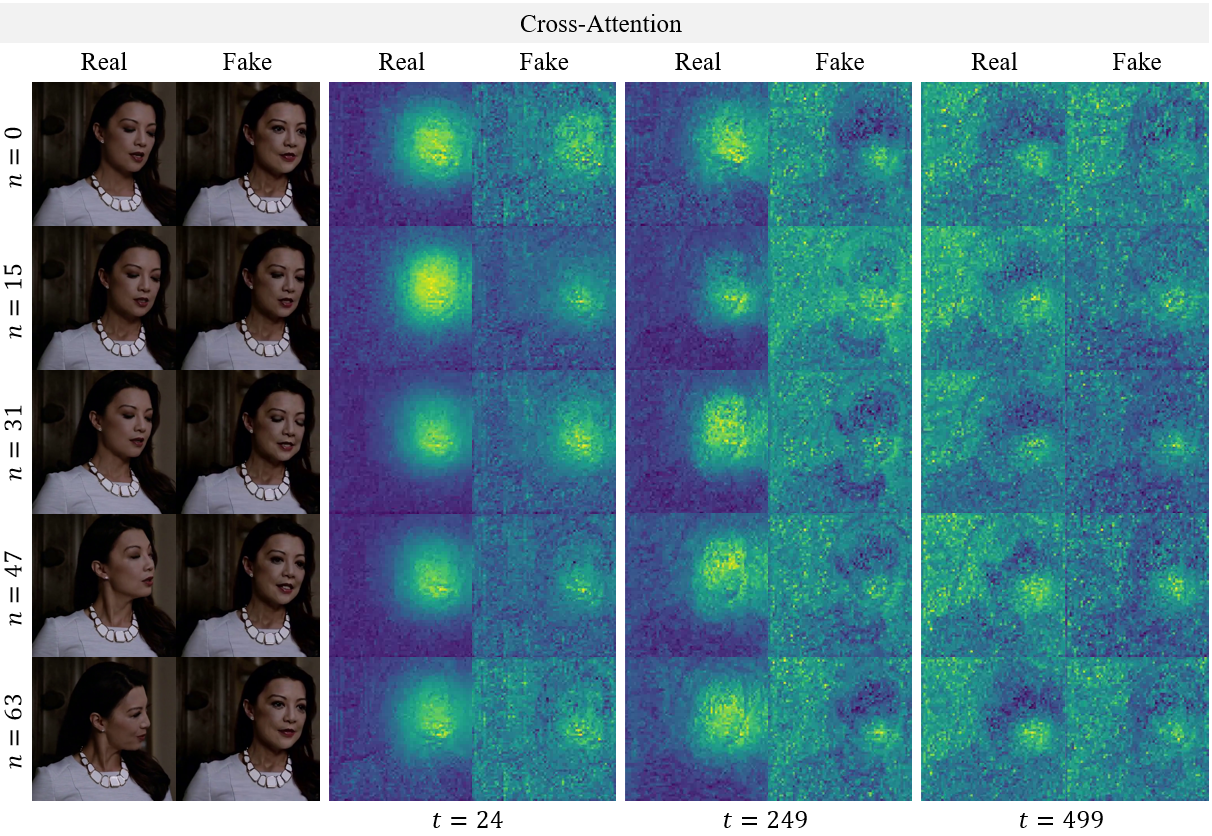} \vspace{0.5cm}
 
 \includegraphics[width=0.9\linewidth]{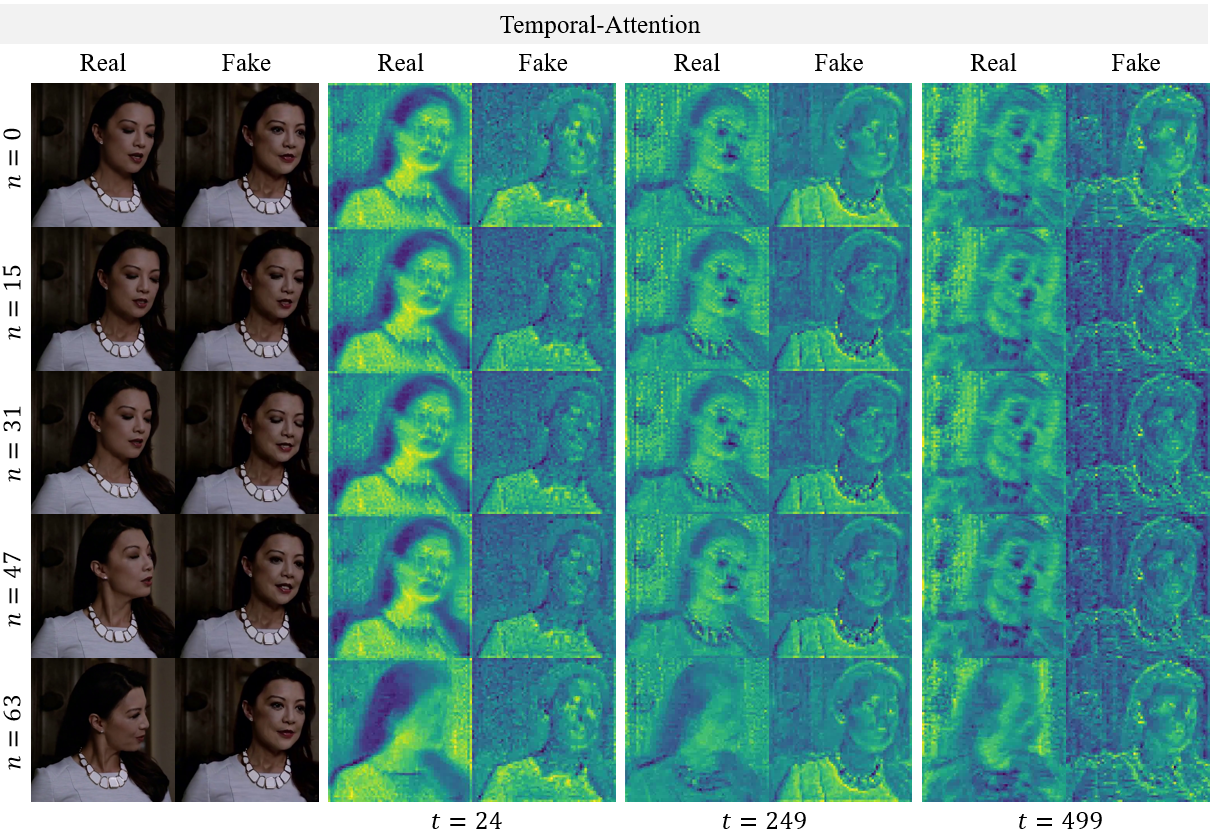}
 \end{center}
\end{figure*}

\begin{figure*}
\setcounter{figure}{0}
\renewcommand{\thefigure}{C.3}
 \begin{center}
 \vspace{-10mm}
 \includegraphics[width=0.9\linewidth]{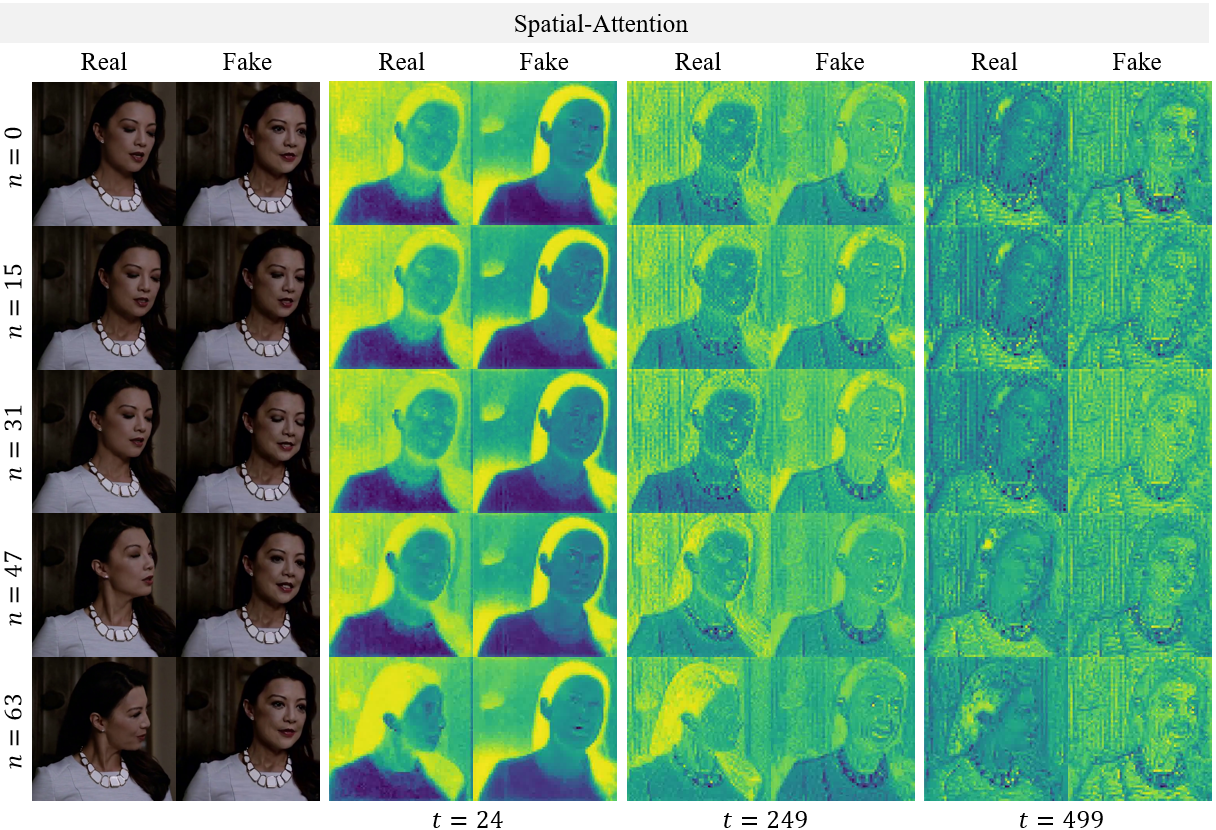}
 \end{center}
  \caption{\textbf{Visualization of attention features across diffusion timesteps.}}
\label{fig:b3_abl_attn}
\end{figure*}

\subsection{Choice of Attention Type and Timestep}
We present additional visual examples across different attention types and DDIM inversion timesteps $t$ in Figure~\ref{fig:b3_abl_attn}. In a diffusion model trained with $T=1000$ steps, we perform inversion with a 40 step sampling schedule and conducted an ablation study over cross-attention, spatial-attention, and temporal-attention, comparing three representative timesteps at $t\in\{24,249,499\}$. Audio-visual cross-attention consistently concentrated on articulators (e.g., lips, jaw), while suppressing the background. Furthermore, as $t$ increased, cross-attention maintained the highest performance at every timestep, outperforming both temporal and spatial attention, indicating that it is the most robust component (see Table~\ref{tab:6_ablation_attn_feat}).

\noindent \textbf{Analysis on Chosen Timestep.} As reported in Table~\ref{tab:6_ablation_attn_feat} of the main paper, the performance of X-AVDT consistently improved as the timestep decreased, across cross-attention, temporal-attention, and spatial-attention. This tendency likely arises because features become more informative in earlier diffusion steps (i.e., as $t \rightarrow 0$), while features in later steps are more heavily corrupted by noise and thus less discriminative. This observation aligns with prior findings~\cite{tang2023emergent,zhang2023tale,yun2024representative,luo2023diffusion,lee2025stylemm,stracke2025cleandift}, which show that mid-to-early diffusion features provide stronger signals for correspondence, stylization, and segmentation due to reduced noise perturbation and richer structural detail. Therefore we did not conducted experiments on $t > 500$, following the previous research. 

%% file: sup_sec/D_Analysis.tex
\section{Analysis}
\label{sup:d}

\begin{figure}[t]
\setcounter{figure}{0}
\renewcommand{\thefigure}{D.1}
 \includegraphics[width=\columnwidth]{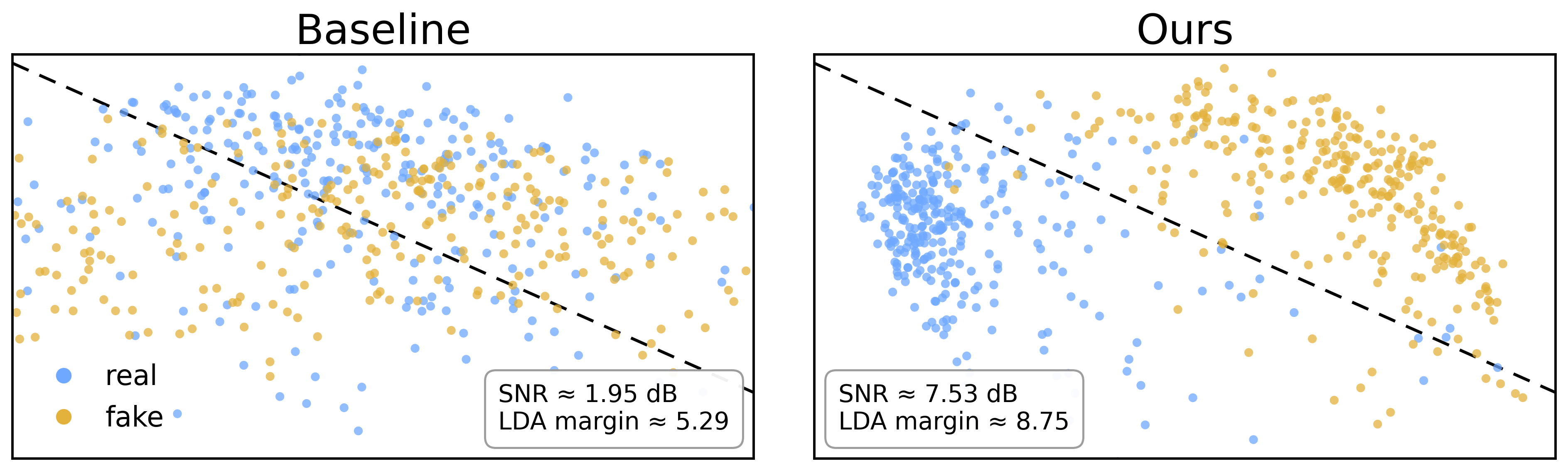}
 \caption{\textbf{PCA embeddings with a shared LDA decision boundary.} Embeddings are extracted from the baseline and from the independently trained X-AVDT, without any fine-tuning.} 
 \label{fig:snr}
\end{figure}

This section complements our method by analyzing the overall detection pipeline, and visualizing the internal audio-visual cross-attention maps from the diffusion backbone. We compared our method against a visual-only baseline~\cite{cazenavette2024fakeinversion}. For the visualization, we present attention heatmaps for the source and representative generators.

\subsection{Fisher SNR and LDA Margin}
We hypothesize that internal audio-visual cross-attention features from large generative models provide a strong discriminative signal for deepfake detection. As shown in Figure \ref{fig:snr}, our method produced noticeably better real/fake separation than a visual only baseline \cite{cazenavette2024fakeinversion}. For a fair comparison, we fit a single linear discriminant analysis (LDA) classifier in the embedding space and project both methods to two dimensions, using the same decision boundary across panels. Measured by Fisher signal-to-noise ratio (SNR) \cite{fisher1936use} and the LDA margin, performance improves from \textbf{1.95dB and 5.29} (baseline) to \textbf{7.53dB and 8.75} (ours). This gap suggests that cross-attention captures stable audio-visual correspondence and exposes inconsistencies that generative models fail to reproduce.

\begin{figure*}
\setcounter{figure}{0}
\renewcommand{\thefigure}{D.2}
\centering
\vspace{-1cm}
\includegraphics[width=0.8\linewidth]{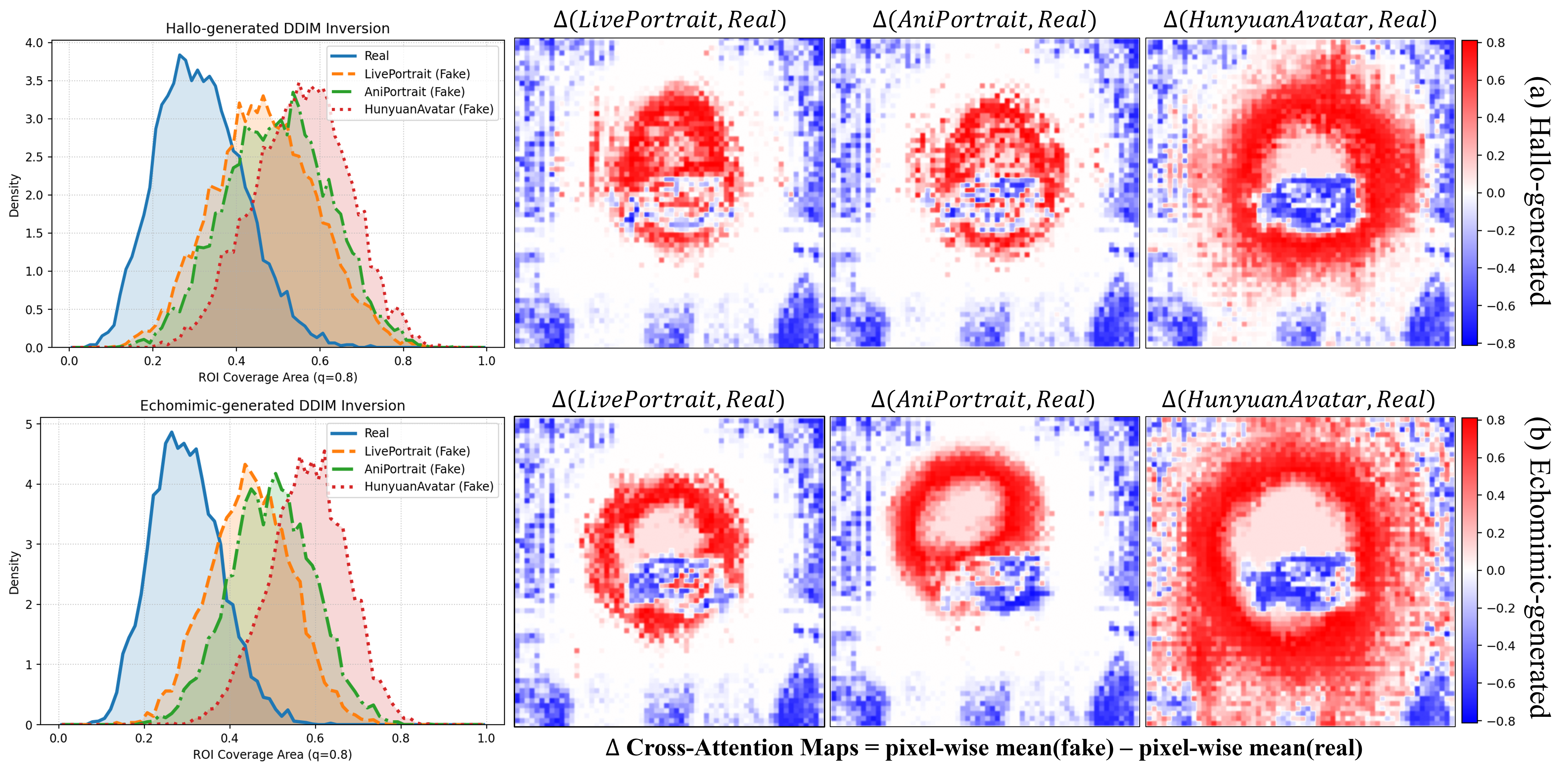}
\caption{\textbf{Top-$q$ Attention Mass Coverage within the Face ROI (Left) and $\Delta$ Cross-Attention Maps (Right).} In the $\Delta$ maps, red indicates regions with higher fake cross-attention than real ($\Delta>0$), and blue indicates the opposite ($\Delta<0$).}
\label{fig:c2_attnention_robustness}
\end{figure*}

\begin{figure}[h]
\setcounter{figure}{0}
\renewcommand{\thefigure}{D.3}
\includegraphics[width=\columnwidth]{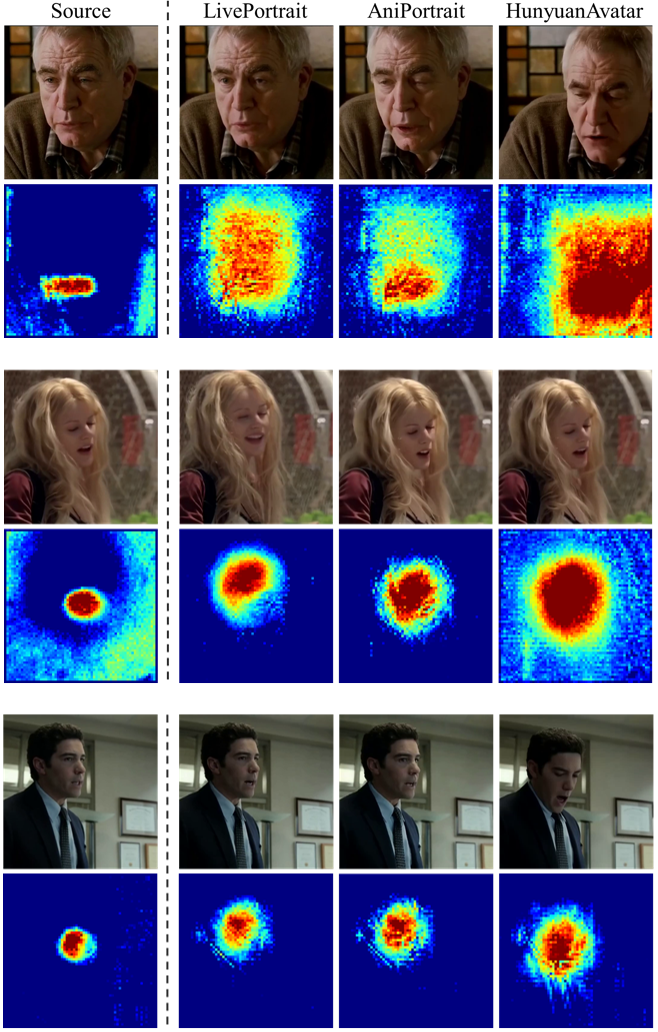}
\caption{\textbf{Temporally averaged cross-attention heatmaps.}}
\label{fig:c3_av_attn_map}
\end{figure}

\subsection{Cross-Attention Robustness}
Figure~\ref{fig:c2_attnention_robustness} quantifies the top-$q$ attention mass coverage within the face ROI: real videos concentrate the top-$q$ mass in a smaller ROI, whereas synthesized videos consistently require coverage of a larger ROI coverage (left). Moreover, the $\Delta$ attention maps reveal a coherent spatial contrast pattern: attention for real videos is concentrated on the mouth and background, while attention for fake videos is more broadly distributed along the face boundary (right). This pattern persists across two different inversion sources, Hallo~\cite{xu2024hallo}, our backbone generator, and Echomimic~\cite{chen2025echomimic}.

\subsection{Attention Map Visualization}
To complement our quantitative results, we visualize internal audio-visual cross-attention maps from the diffusion backbone. As shown in Figure~\ref{fig:c3_av_attn_map}, for each video we extract cross-attention during DDIM inversion, normalize the weights per frame, and average them over time to obtain a single heatmap. We compare the source clip deepfake results from with three representative generators that span different synthesis frameworks: LivePortrait (GAN-based)~\cite{guo2024liveportrait}, AniPortrait (diffusion-based)~\cite{wei2024aniportrait}, and HunyuanAvatar (flow-matching-based)~\cite{chen2025hunyuanvideo}. Empirically, similar cross-attention patterns are observed across the results from different generator frameworks, indicating that large generative models already provide strong and efficient self-supervised representations well suited for detection. Note that our detector is trained and evaluated on attention features, not on visualized maps, which are provided solely for interpretability. Averaging and normalization for display can introduce information loss, whereas feature vectors are stable.

%% file: sup_sec/E_dataset.tex
\vspace{-1mm}
\section{MMDF Dataset}
\label{sup:e}
\subsection{MMDF Construction and Split Protocol}
The filtering in MMDF construction corresponds to standard face-detection preprocessing~\cite{lugaresi2019mediapipe} that is widely adopted across facial video datasets and generation pipelines. We only removed clips with inaccurate face tracking to ensure reliable ground-truth pairs, thereby reducing confounding failure cases across all compared detectors rather than favoring X-AVDT. This reduced the candidate set by 6.93\% (2,001 clips removed). MMDF is intentionally designed as a strict cross-generator generalization benchmark. Because our goal is to detect forgeries from unseen generation mechanisms, we enforce disjoint generation methods between train and test to avoid overfitting to generator-specific artifacts. We also incorporate a variety of generators, model families, and synthesis methods to maximize the diversity of train-test combinations under this cross-setting.

\subsection{Details of Fake Generators}
As mentioned in the main paper, we adopt the Hallo3 dataset released by its authors~\cite{cui2025hallo3} as the source corpus and employ a curated subset as our real set (see Figure~\ref{fig:d1_dataset_distribution}). Then the generators described below synthesize the paired fakes. During preprocessing, all videos are sampled at 25fps to obtain the reference images, and the generated sequences are temporally aligned to their sources for one-to-one pairing, and resized to $512\times512$. All fakes are produced by inference only, without any additional training, using the authors’ default parameters.

\subsubsection{\texorpdfstring{Hallo2~\cite{cui2024hallo2}}{Hallo2}}
\vspace{-1mm}
Hallo2 is a diffusion-based, audio-driven portrait image animation model. For the fake samples in the MMDF training set, we use the first video frame as the single reference image and feed the clip's corresponding audio as the driving signal to generate an audio-synchronized talking-head sequence.

\subsubsection{\texorpdfstring{LivePortrait~\cite{guo2024liveportrait}}{LivePortrait}}
\vspace{-1mm}
LivePortrait is a GAN-based portrait animation method that warps a single reference image according to a driving signal to perform self-reenactment. For the fake samples in the MMDF training set, we use the first frame as the reference image and animate it using the remaining frames as driving images. Note that LivePortrait operates as an image-to-video model without audio driving (i.e., motion is driven solely by image frames).

\subsubsection{\texorpdfstring{FaceAdapter~\cite{han2024face}}{FaceAdapter}}
\vspace{-1mm}
FaceAdapter is a face-editing adapter for pretrained diffusion models, targeting face swapping. For the fake samples in the MMDF training set, we randomly select a source identity and a target identity. Leveraging its image-to-image design, we generate swapped frames for the target clip and then pair the synthesized frames with the source audio to produce the final video, thereby preserving the source identity and speech.

\begin{figure}[t]
\vspace{-4mm}
\setcounter{figure}{0}
\renewcommand{\thefigure}{E.1}
\includegraphics[width=\linewidth]{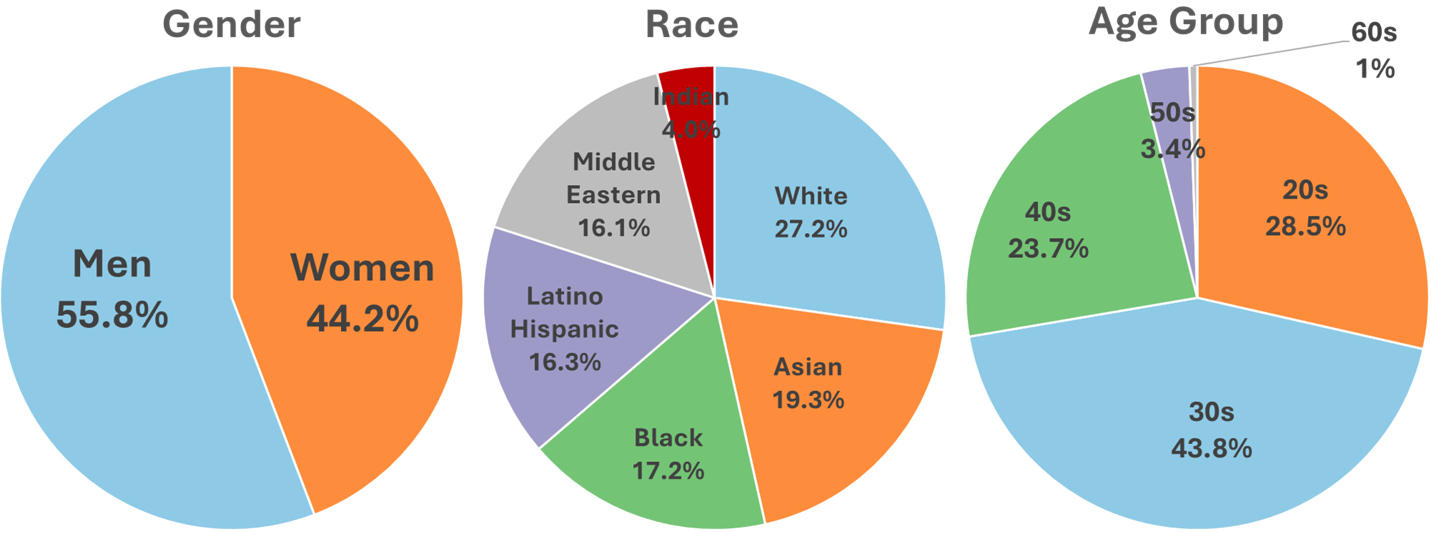}
\caption{\textbf{Statistics of the MMDF dataset.}}
\label{fig:d1_dataset_distribution}
\end{figure}

\subsubsection{\texorpdfstring{HunyuanAvatar~\cite{chen2025hunyuanvideo}}{HunyuanAvatar}}
\vspace{-1mm}
HunyuanAvatar is a flow-matching-based audio-driven human animation method. For the fake samples in the MMDF evaluation set, we use the first frame as the source input. We then feed the clip's corresponding audio together with a text prompt generated from the first frame by BLIP-2, OPT-2.7b model, producing an audio-synchronized, text-conditioned talking head sequence.

\subsubsection{\texorpdfstring{MegActor-$\Sigma$~\cite{yang2025megactor}}{MegActor-$\Sigma$}}
\vspace{-1mm}
MegActor-$\Sigma$ is a diffusion-transformer (DiT)–based portrait animation method with mixed-modal conditioning. For the fake samples in the MMDF evaluation set, we use the first frame as the source input, and feed the source video together with its corresponding audio to generate a self-reenactment sequence.

\subsubsection{\texorpdfstring{Aniportrait~\cite{wei2024aniportrait}}{Aniportrait}}
\vspace{-1mm}
Aniportrait is a diffusion-based, audio-driven portrait animation method. For the fake samples in the MMDF evaluation set, we use the first video frame as the single reference image and feed the clip's corresponding audio as the driving signal to generate an audio-synchronized talking-head sequence.

\subsection{Input Representation Visualization}
Figure~\ref{fig:d2_input1} and Figure~\ref{fig:d2_input2} contain samples from our model's input representation, video composite $\boldsymbol{\psi}$ and AV cross-attention feature $\boldsymbol{\psi}$ utilized by our detector. From top to bottom, we show the audio-visual cross-attention feature $\boldsymbol{\psi}$, the original video $x$, the decoded latent DDIM noise map $D(\hat z_T)$, the reconstructed video $D(\hat z_0)$, and the reconstruction residual $r=\lvert x-D(\hat z_0)\rvert$ of the video composite $\boldsymbol{\phi}$. The supplementary video with audio further illustrates temporal dynamics and audio-visual synchronization patterns.

\subsection{MMDF Dataset Visualization}
Figures~\ref{fig:d3_1_hallo2}--\ref{fig:d3_6_aniportrait} present samples from the curated MMDF dataset used in our experiments. For each identity, the real frames were taken from the source dataset~\cite{cui2025hallo3} and the fake videos were generated by respective generators. The supplementary video with audio further illustrates temporal dynamics and audio-visual synchronization patterns.

\section{Limitations}
\label{sup:f}

\renewcommand{\thefigure}{\Alph{figure}}
\setcounter{figure}{5} 
\begin{figure}[h]
\vspace{-2mm}
    \centering
    \includegraphics[width=0.7\linewidth]{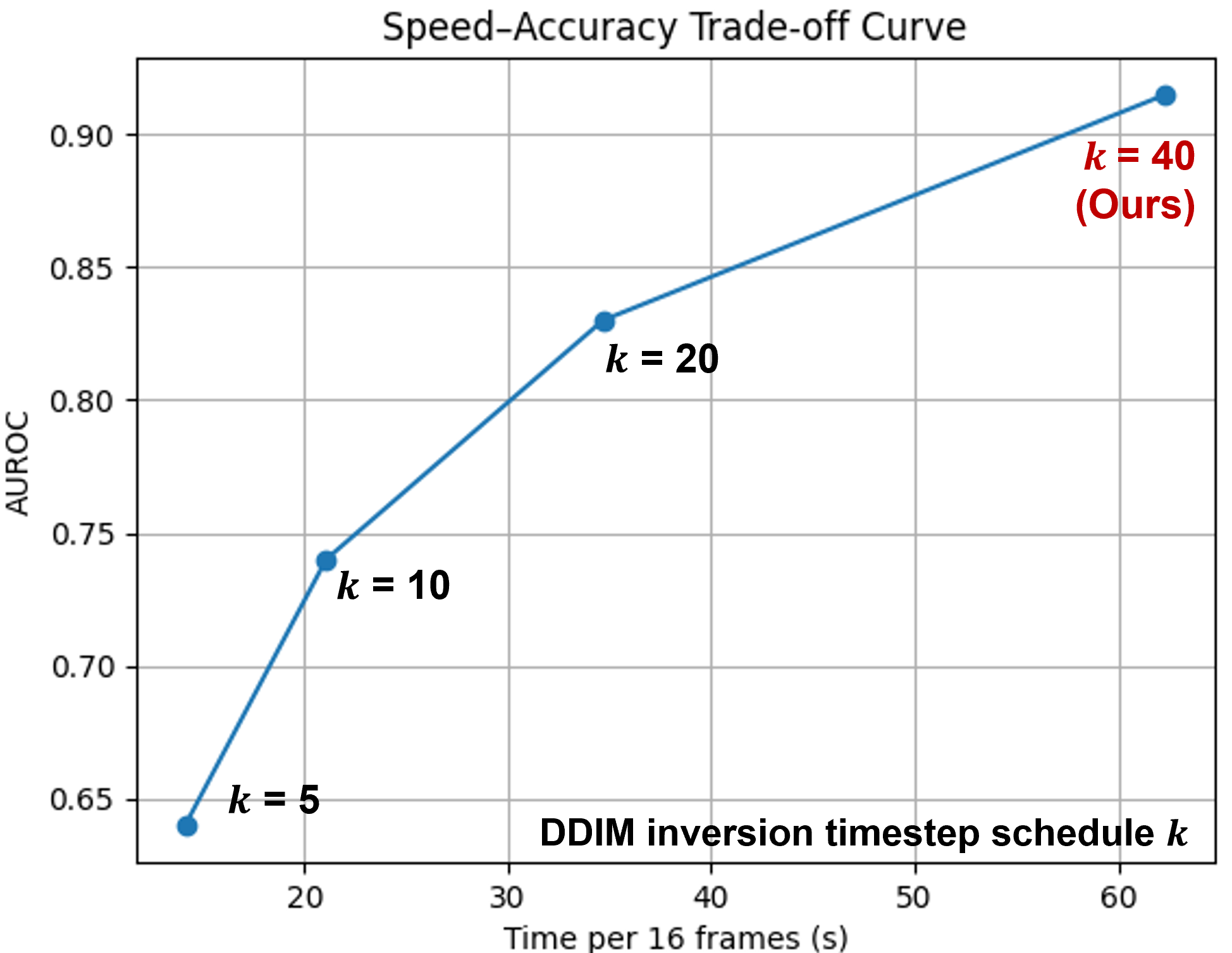}
    \caption{\textbf{Speed-accuracy trade-off with DDIM inversion steps.}}
    \label{fig:tradeoff}
\end{figure}

\noindent Figure~\ref{fig:tradeoff} indicates a potential limitation of our approach in real-world applications. Performance depends on the number of DDIM inversion timestep schedule $k$ used to extract $\boldsymbol{\phi}$ and $\boldsymbol{\psi}$. While larger $k$ yields more faithful inversion features and improves AUROC, it increases runtime and computational cost. Conversely, smaller $k$ reduces latency but degrades detection accuracy. The incurred cost can be mitigated in future work by adopting fewer step schedules or model distillation.

\begin{figure*}[t]
\setcounter{figure}{0}
\renewcommand{\thefigure}{D.2.1}
\vspace{-0.7cm}
\centering
\includegraphics[height=1\textheight]{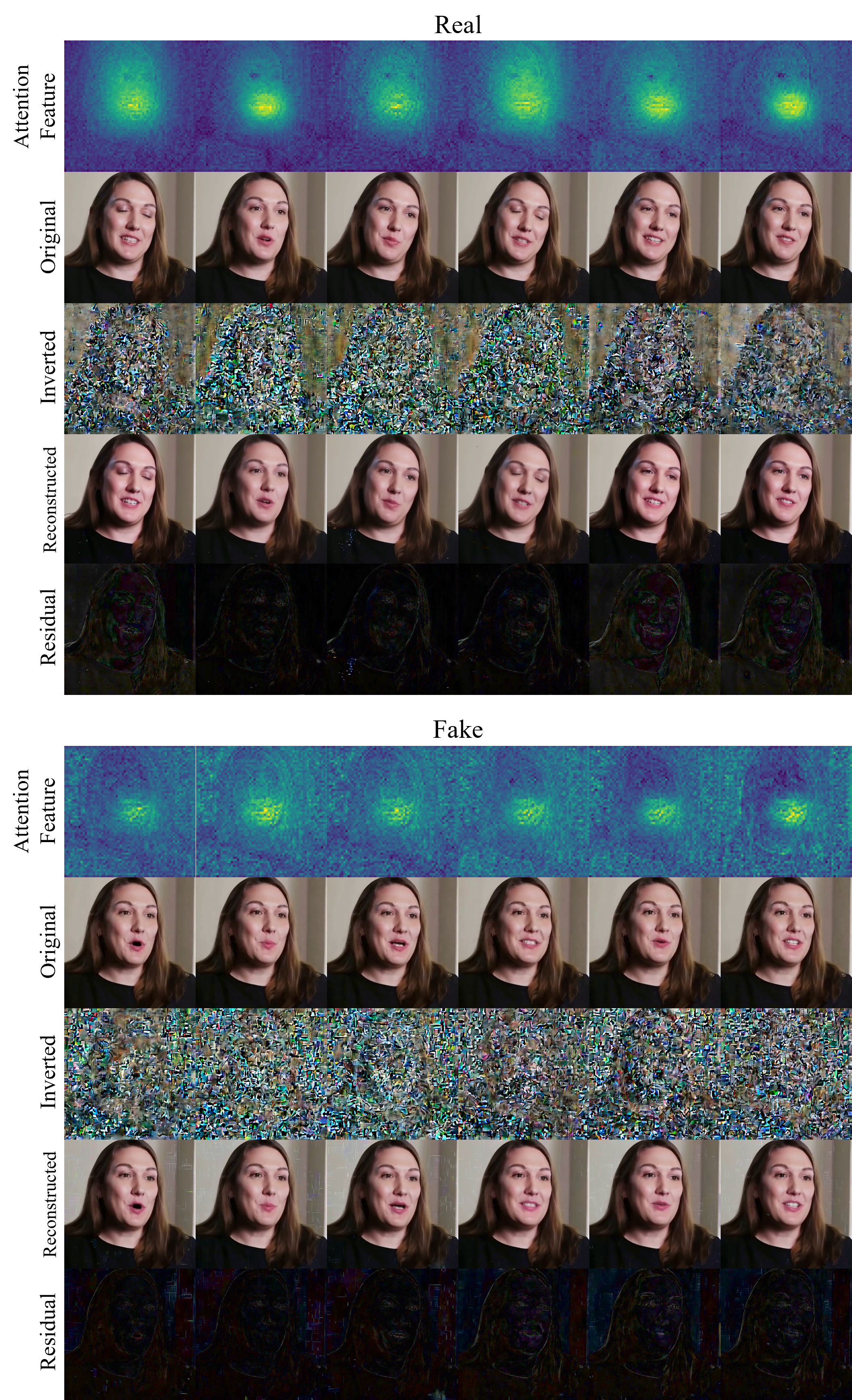}
\caption{\textbf{Qualitative visualization of the input representations $\boldsymbol{\phi}$ and $\boldsymbol{\psi}$.}}
\label{fig:d2_input1}
\end{figure*}

\begin{figure*}[t]
\setcounter{figure}{0}
\renewcommand{\thefigure}{D.2.2}
\vspace{-0.7cm}
\centering
\includegraphics[height=1\textheight]{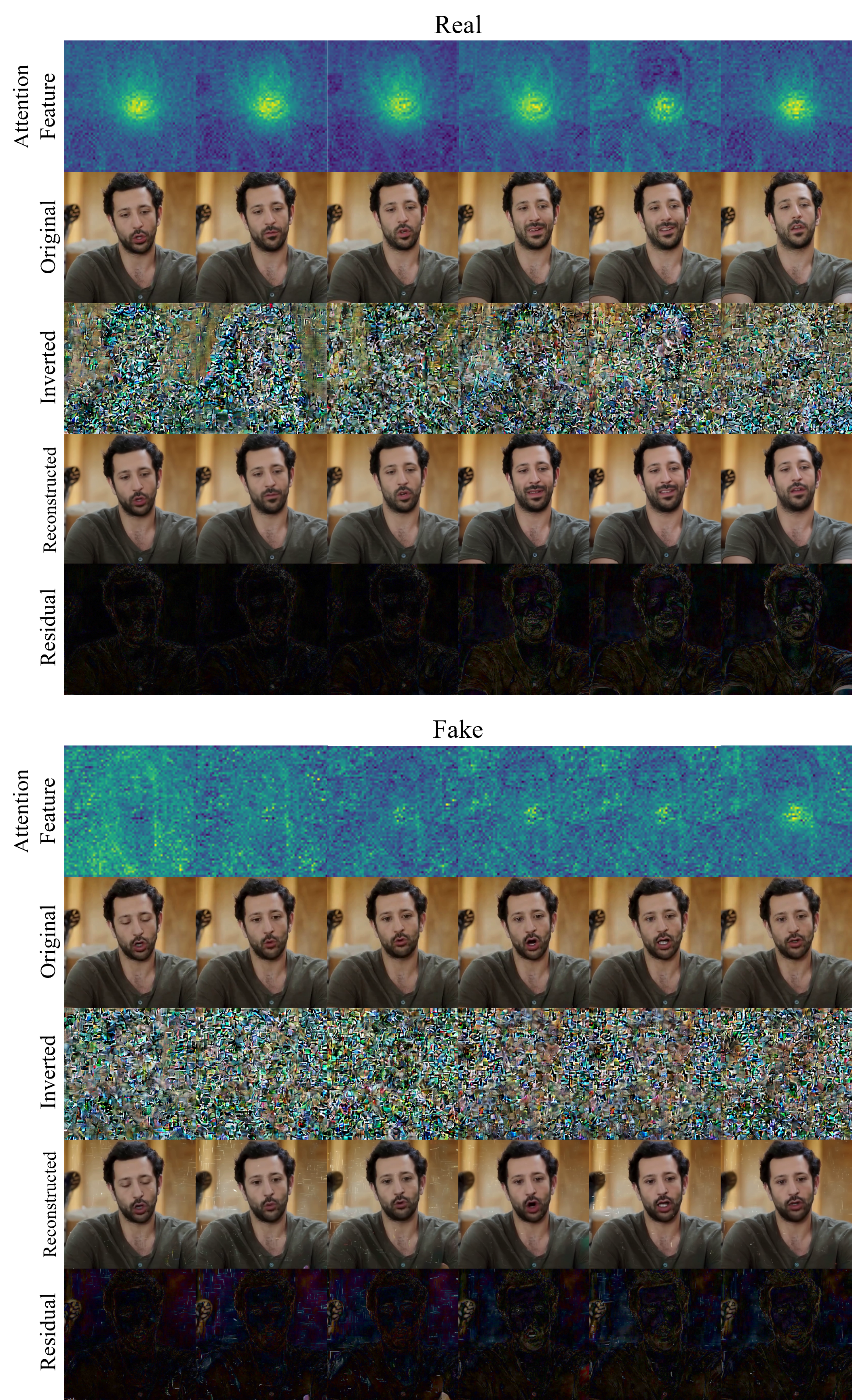}
\caption{\textbf{Qualitative visualization of the input representations $\boldsymbol{\phi}$ and $\boldsymbol{\psi}$.}}
\label{fig:d2_input2}
\end{figure*}

\begin{figure*}[t]
\setcounter{figure}{0}
\renewcommand{\thefigure}{D.3.1}
\centering
\includegraphics[width=0.9\linewidth]{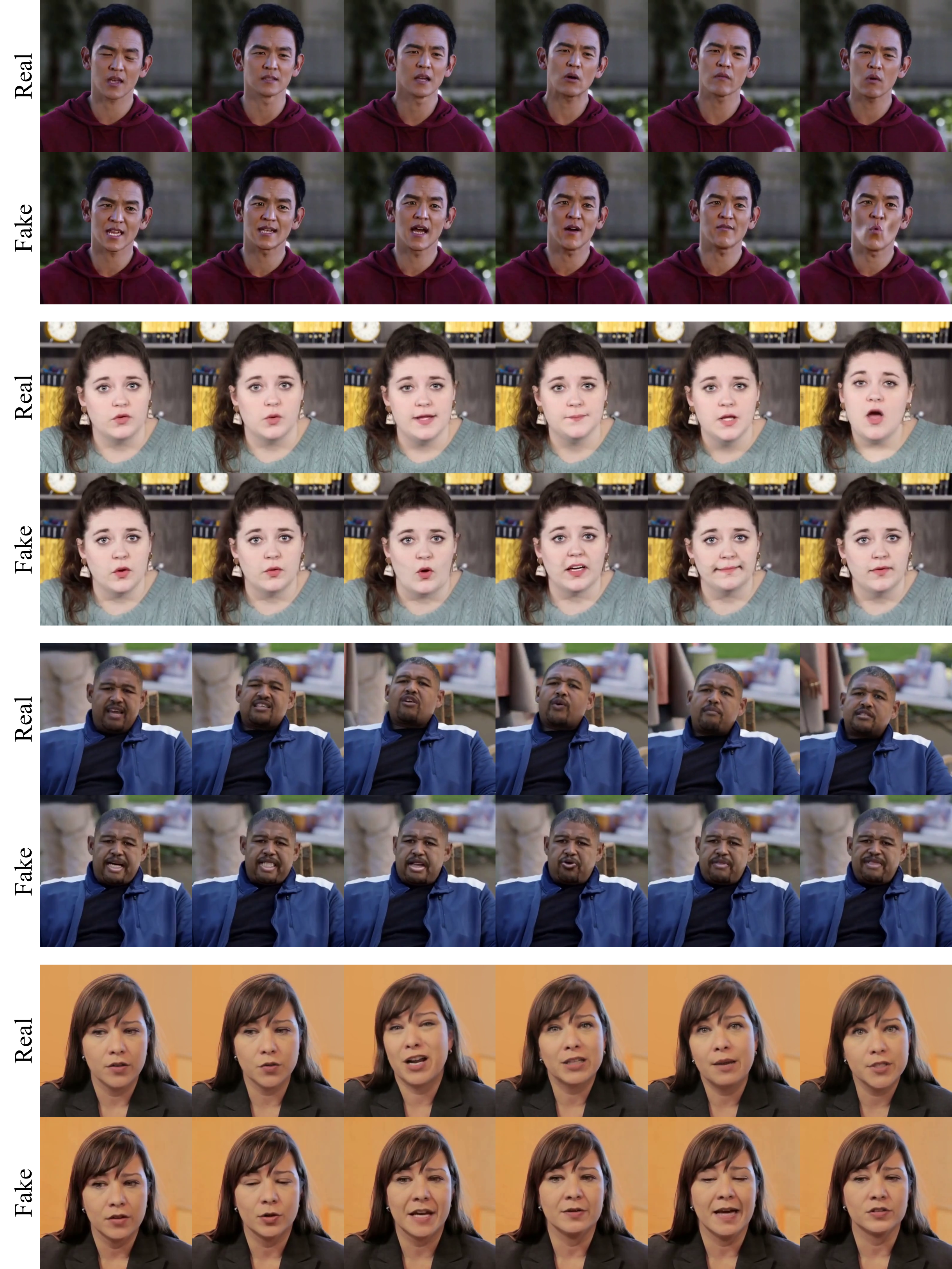}
\caption{\textbf{Qualitative comparison of real and fake videos generated by Hallo2~\cite{cui2024hallo2} in the MMDF.}}
\label{fig:d3_1_hallo2}
\end{figure*}

\begin{figure*}[t]
\setcounter{figure}{0}
\renewcommand{\thefigure}{D.3.2}
\centering
\includegraphics[width=0.9\linewidth]{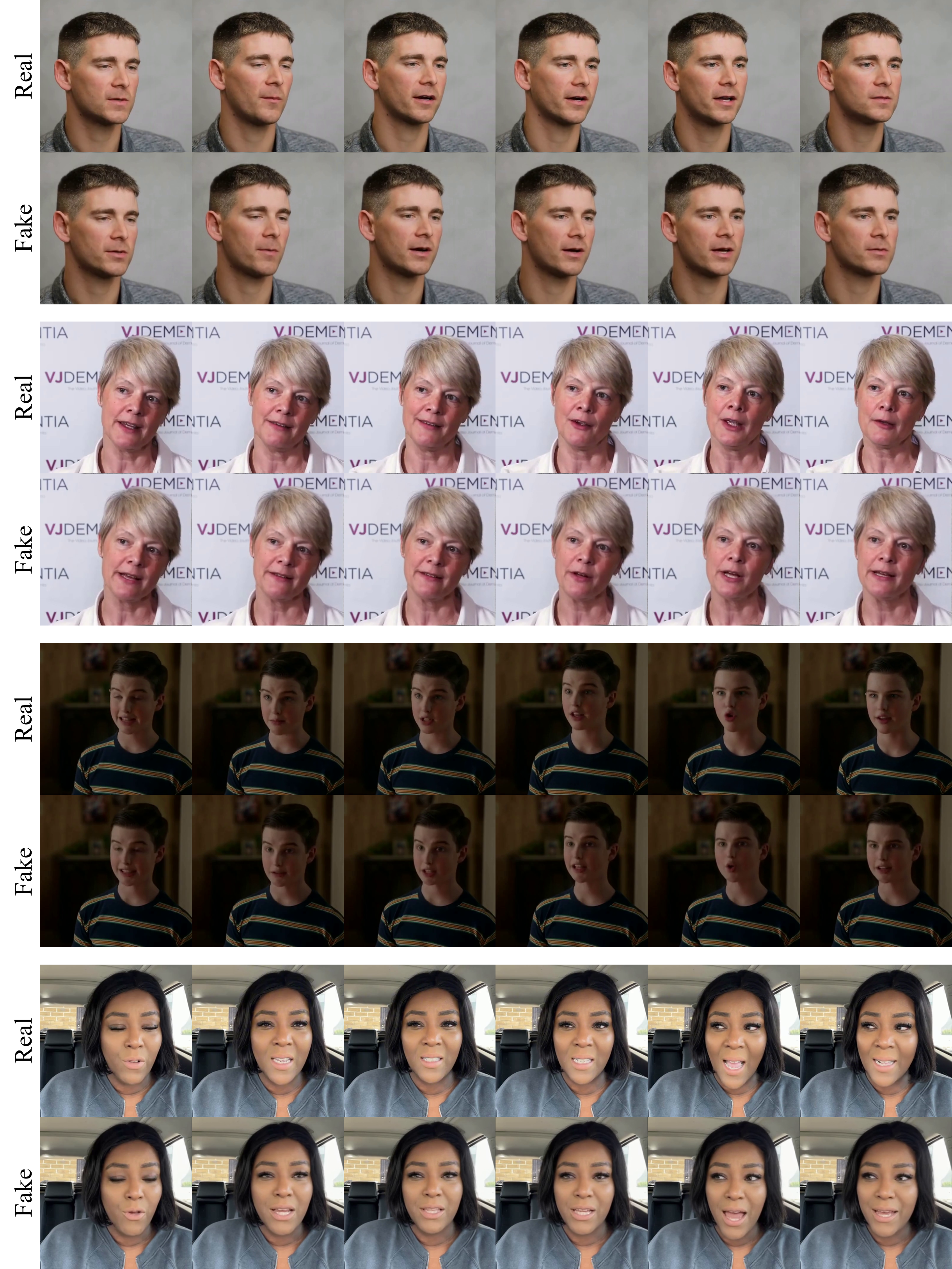}
\caption{\textbf{Qualitative comparison of real and fake videos generated by LivePortrait~\cite{guo2024liveportrait} in the MMDF.}}
\label{fig:d3_2_liveportrait}
\end{figure*}

\begin{figure*}[t]
\setcounter{figure}{0}
\renewcommand{\thefigure}{D.3.3}
\centering
\includegraphics[width=0.9\linewidth]{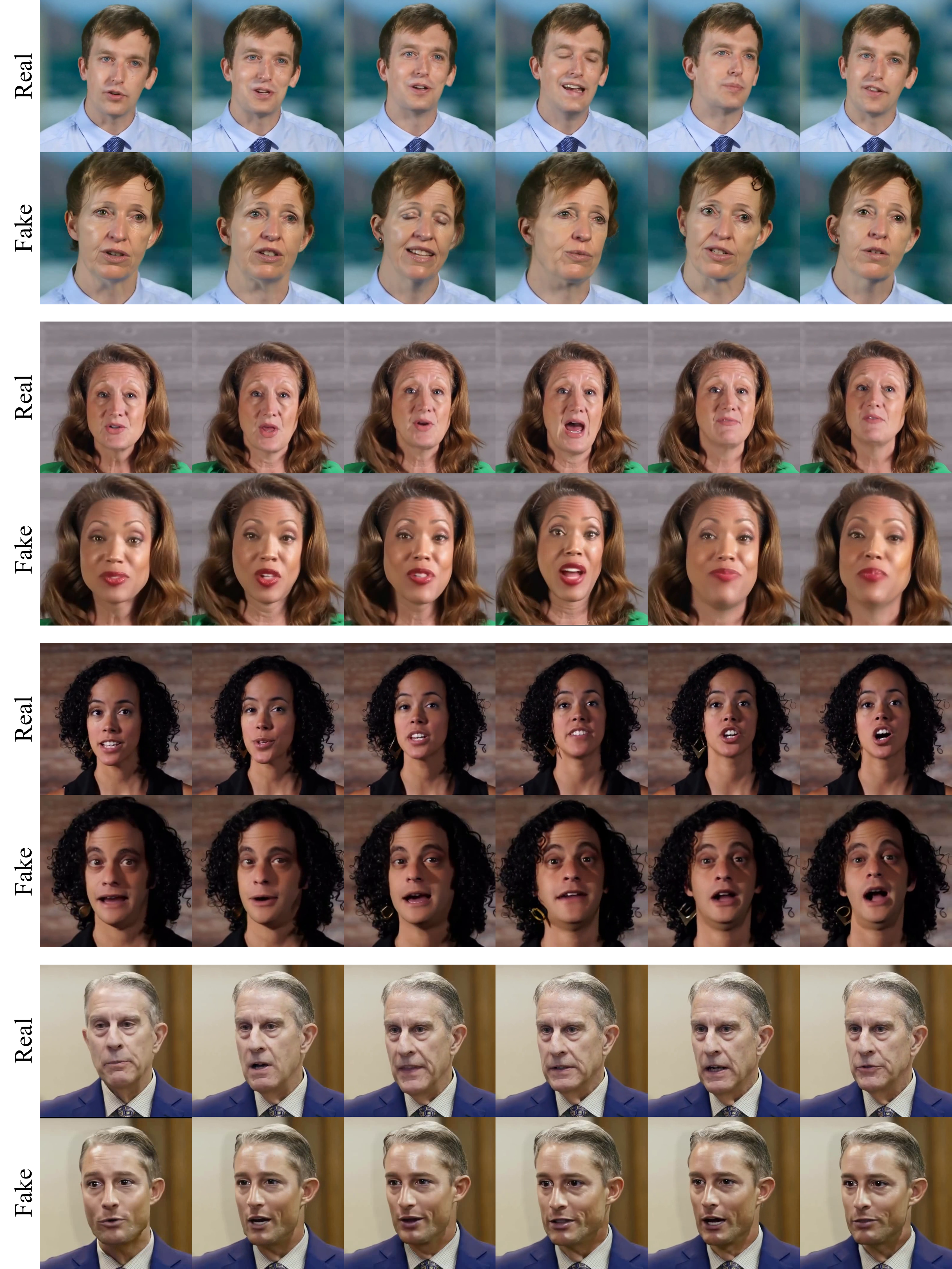}
\caption{\textbf{Qualitative comparison of real and fake videos generated by FaceAdater~\cite{han2024face} in the MMDF.}}
\label{fig:d3_3_faceadatper}
\end{figure*}

\begin{figure*}[t]
\setcounter{figure}{0}
\renewcommand{\thefigure}{D.3.4}
\centering
\includegraphics[width=0.9\linewidth]{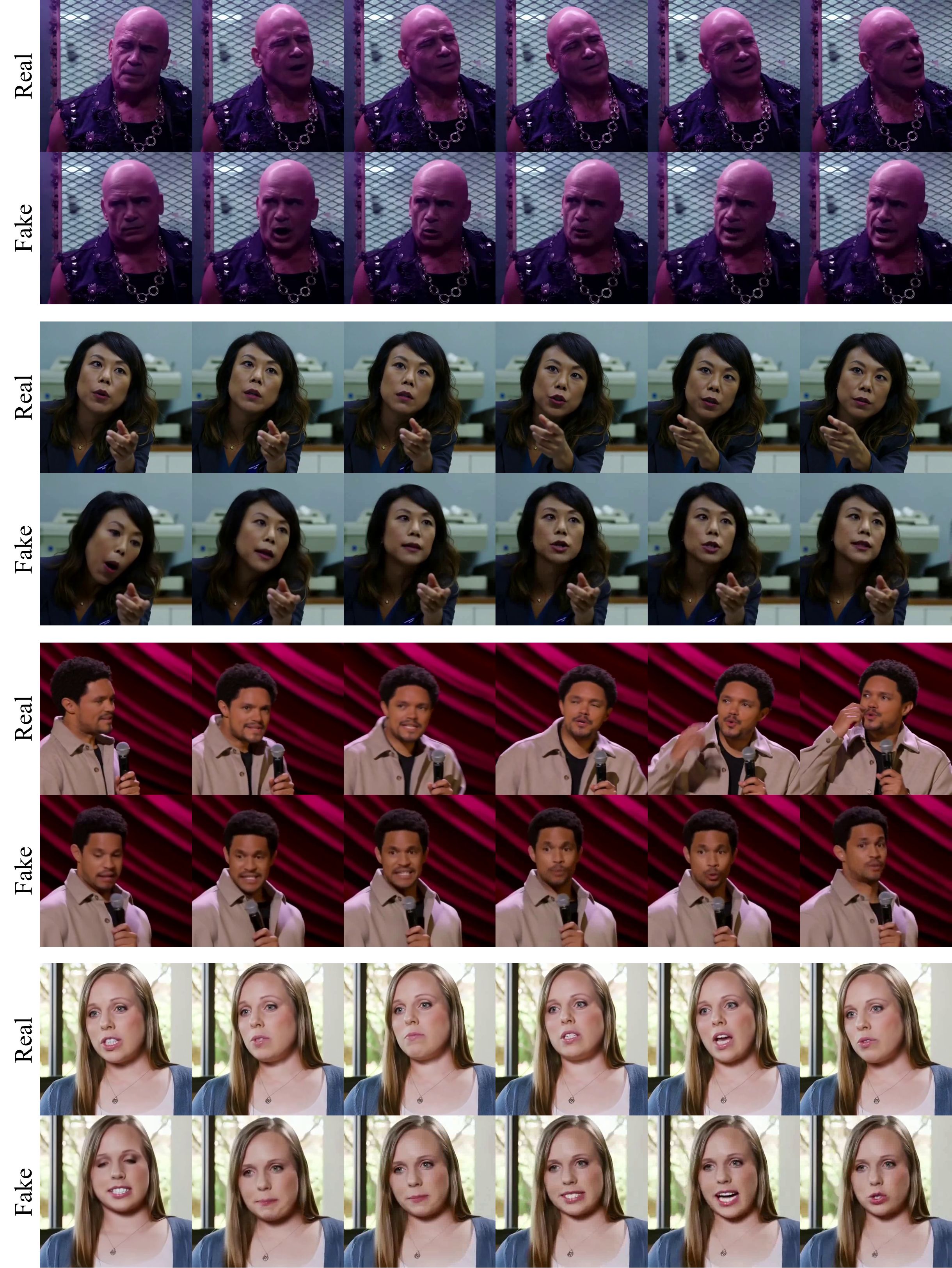}
\caption{\textbf{Qualitative comparison of real and fake videos generated by HunyuanAvatar~\cite{chen2025hunyuanvideo} in the MMDF.}}
\label{fig:d3_4_hunyuan}
\end{figure*}

\begin{figure*}[t]
\setcounter{figure}{0}
\renewcommand{\thefigure}{D.3.5}
\centering
\includegraphics[width=0.9\linewidth]{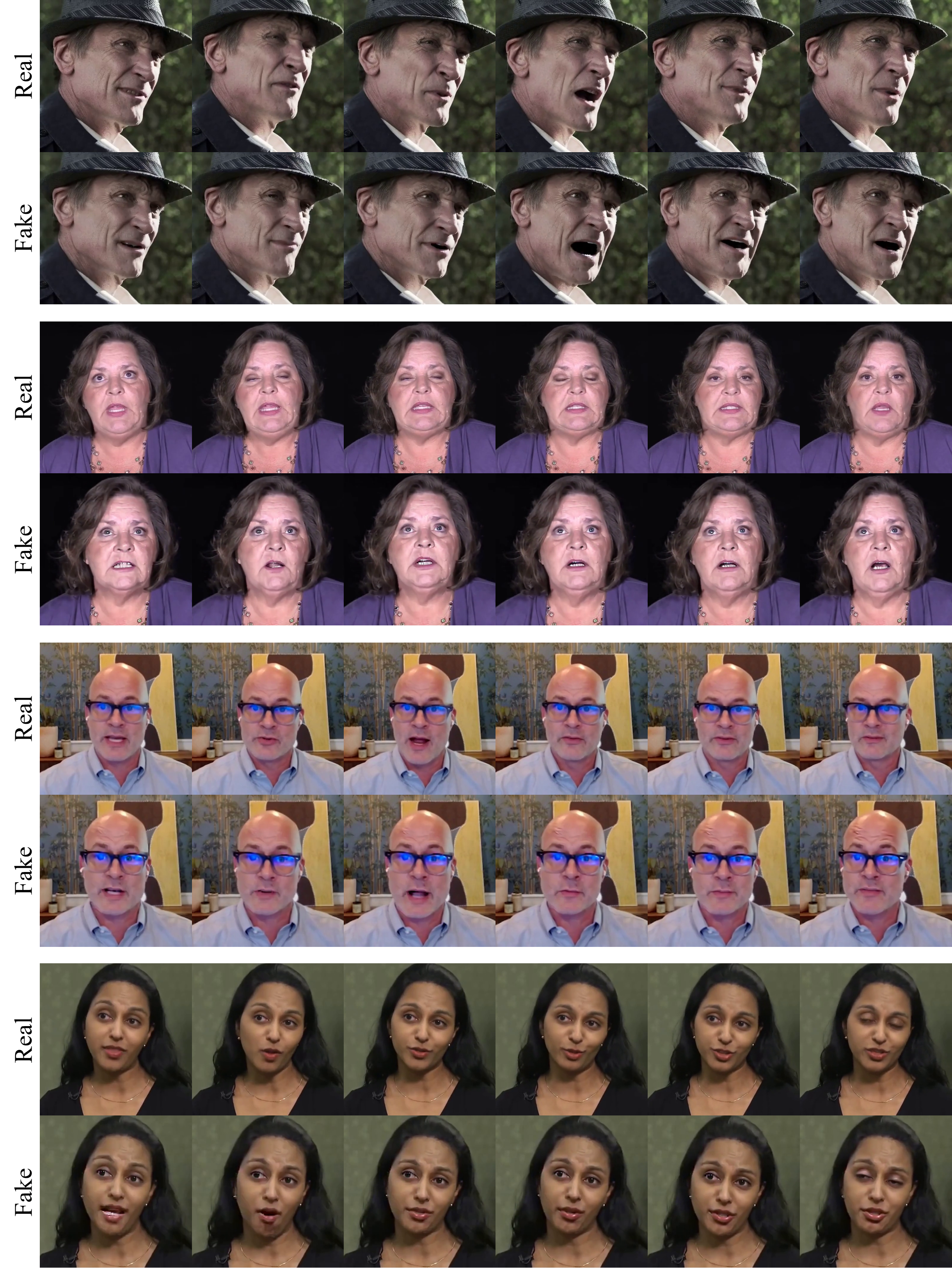}
\caption{\textbf{Qualitative comparison of real and fake videos generated by MegActor-$\Sigma$~\cite{yang2025megactor} in the MMDF.}}
\label{fig:d3_5_megactor}
\end{figure*}

\begin{figure*}[t]
\setcounter{figure}{0}
\renewcommand{\thefigure}{D.3.6}
\centering
\includegraphics[width=0.9\linewidth]{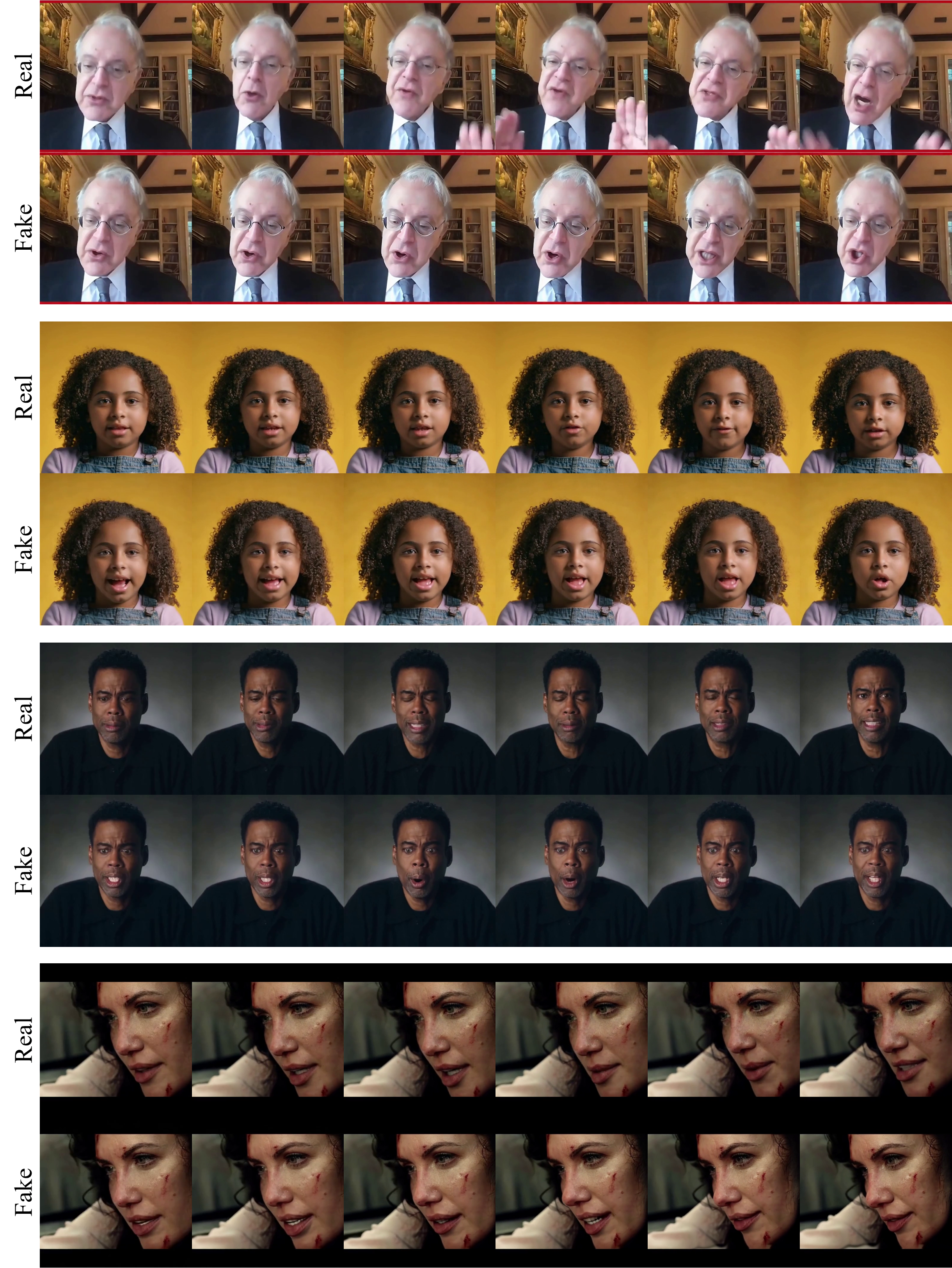}
\caption{\textbf{Qualitative comparison of real and fake videos generated by AniPortrait~\cite{wei2024aniportrait} in the MMDF.}}
\label{fig:d3_6_aniportrait}
\end{figure*}